\documentclass[
  journal=nlp,
  manuscript=article,
  year=2025,
  volume=1,
]{cup-journal}

\usepackage{graphicx}
\usepackage{amsmath}
\usepackage{booktabs}
\usepackage{setspace}
\usepackage{todonotes}
\usepackage{multirow}
\usepackage{makecell}
\usepackage{datenumber}
\usepackage[nopatch]{microtype}
\usepackage{skak}

\usepackage{tikz}
\usetikzlibrary{positioning, shapes, calc, fit, decorations.pathmorphing, decorations.shapes}
\tikzset{>=latex}

\usepackage{pgf} % for calculating the values for gradient
\usetikzlibrary{backgrounds}

\newcommand{\badgeOneTower}{%
  \smash{\begin{tikzpicture}[baseline=-0.8ex]
  \node[font=\sffamily\scriptsize, draw, line width=0.4pt, inner sep=2pt] {1\symrook E};
  \end{tikzpicture}}%
}

\newcommand{\badgeTwoTower}{%
  \smash{\begin{tikzpicture}[baseline=-0.8ex]
  \node[font=\sffamily\scriptsize, draw, line width=0.4pt, inner sep=2pt] {2\symrook E};
  \end{tikzpicture}}%
}

\newcommand{\badgeComboEncoder}{%
  \smash{\begin{tikzpicture}[baseline=-0.8ex]
  \node[font=\sffamily\scriptsize, draw, line width=0.4pt, inner sep=2pt] {\symrook $\bullet$\symrook};
  \end{tikzpicture}}%
}

\newcommand{\badgeDualEncoder}{%
  \smash{\begin{tikzpicture}[baseline=-0.8ex]
  \node[font=\sffamily\scriptsize, draw, line width=0.4pt, inner sep=2pt] (e1) {E};
  \node[font=\sffamily\scriptsize, right=-3pt of e1] (t) {$\bullet$};
  \node[font=\sffamily\scriptsize, draw, line width=0.4pt, inner sep=2pt,right=-3pt of t] (e2) {E};
  \end{tikzpicture}}%
}

\newcommand{\badgeEncoderDecoder}{%
  \smash{\begin{tikzpicture}[baseline=-0.8ex]
  \node[font=\sffamily\scriptsize, draw, line width=0.4pt, inner sep=2pt] (e) {E};
  \node[font=\sffamily\scriptsize, right=-3pt of e] (t) {$\triangleright$};
  \node[font=\sffamily\scriptsize, draw, line width=0.4pt, circle, inner sep=1.1pt, right=-3pt of t] (d) {D};
  \draw[->,gray!50] (d.north) to[out=30, in=-30, looseness=2.7] (d.south);
  \end{tikzpicture}}%
}

\newcommand{\badgeDecoder}{%
  \smash{\begin{tikzpicture}[baseline=-0.8ex]
  \node[font=\sffamily\scriptsize, draw, line width=0.4pt, circle, inner sep=1.5pt] (d) {D};
  \draw[->,gray!50] (d.north) to[out=30, in=-30, looseness=2.8] (d.south);
  \end{tikzpicture}}%
}

\newcommand{\badgeParams}[1]{%
  \smash{\begin{tikzpicture}[baseline=-0.76ex]
    \node[fill=black!60, text=white, rounded corners=2pt, inner sep=2.5pt, font=\footnotesize\sffamily] {#1};
  \end{tikzpicture}}%
}

\newcommand{\badgeDate}[1]{%
  \smash{\begin{tikzpicture}[baseline=0.72ex, inner sep=0pt, outer sep=0pt]
    \path[use as bounding box] (0,0) rectangle (0.88,0.48);
    \draw[fill=black!60, draw=black!70, line width=0.3pt] (0,0.32) rectangle (0.88,0.44);
    \foreach \x in {0.16, 0.36, 0.56, 0.76} {
      \draw[black!60, line width=0.4pt, fill=white] (\x,0.44) circle (0.032);
    }
    \draw[fill=white, draw=black!40, line width=0.3pt] (0,0) rectangle (0.88,0.32);
    \node[font=\scriptsize\sffamily] at (0.44,0.16) {#1};
  \end{tikzpicture}}%
}

\newcommand{\tiers}[6]{%
\begin{tikzpicture}[baseline=0.3ex]
    \pgfmathsetmacro{\maxval}{max(max(max(max(max(#1,#2),#3),#4),#5),#6)}

    \draw[-] (0.1,0) -- (0.7,0);

    \pgfmathsetmacro{\scale}{0.35/ \maxval}
    \draw[fill=black!80] (0.1, 0) rectangle (0.2, {#1 * \scale}); 
    \draw[fill=black!65] (0.2, 0) rectangle (0.3, {#2 * \scale}); 
    \draw[fill=black!50] (0.3, 0) rectangle (0.4, {#3 * \scale}); 
    \draw[fill=black!35] (0.4, 0) rectangle (0.5, {#4 * \scale});
    \draw[fill=black!20] (0.5, 0) rectangle (0.6, {#5 * \scale}); 
    \draw[fill=black!5] (0.6, 0) rectangle (0.7, {#6 * \scale}); 
\end{tikzpicture}
}
\newcommand{\tabcite}[1]{\begin{minipage}{80pt}\centering\setstretch{0.6} \citet{#1}\end{minipage}}
\newcommand{\focuscell}[1]{\begin{minipage}{35pt}\centering\setstretch{0.6}#1\end{minipage}}

\title{Multilingual Vision-Language Models, A Survey}

\author{Andrei-Alexandru Manea}
\affiliation{Faculty of Mathematics and Physics, Charles University, V Holešovičkách 747/2, Prague, Czech Republic}
\email[Andrei-Alexandru Manea]{manea@ufal.mff.cuni.cz}

\author{Jindřich Libovický}
\affiliation{Faculty of Mathematics and Physics, Charles University, V Holešovičkách 747/2, Prague, Czech Republic}
\email[Jindřich Libovický]{libovicky@ufal.mff.cuni.cz}


\begin{document}

\begin{abstract}

This survey examines multilingual vision-language models that process text and images across languages. We review 33 models and 23 benchmarks, spanning encoder-only and generative architectures, and identify a key tension between language neutrality (consistent cross-lingual representations) and cultural awareness (adaptation to cultural contexts). Current training methods can, to some extent, enforce language neutrality, which can be reasonably measured, while cultural awareness depends on diverse data and is difficult to evaluate. Two-thirds of evaluation benchmarks use translation-based approaches prioritizing semantic consistency, though recent work incorporates culturally grounded content. This indicates a discrepancy between training objectives and evaluation goals.

\end{abstract}

\section{Introduction}

Vision-language models are machine learning systems designed for the joint processing of natural language and visual data, enabling tasks such as visual question answering or image retrieval based on text. Building upon recent advances in language modeling and computer vision, this class of models has made unprecedented progress. However, most of the model capabilities are showcased and evaluated in English. There is relatively less work focusing on other, often less-resourced languages, where performance significantly lags behind English and other well-supported languages. Still, there are researchers who put significant effort into making vision-language models accessible worldwide, both by developing multilingual models and by creating benchmarks that enable more accurate evaluation of their performance across languages. In this article, we provide a comprehensive survey of both multilingual vision-language models and multilingual benchmarks, identify and categorize the main challenges in this area, and discuss how models and benchmarks address these challenges.
We focus exclusively on models that process static images alongside language, excluding those designed for videos or for generating images from text.

Vision-language models closely follow developments in language modeling. Historically, many advancements in language modeling were first developed for English (e.g., the original masked-language model BERT, \cite{devlin-etal-2019-bert}, in late 2018), with their multilingual adaptations and vision-language adaptations (e.g., LXMERT, \cite{tan-bansal-2019-lxmert}, in the summer of 2019) occurring later. Multilingual extensions of vision-language models remain even further behind (M3P \cite{ni_m3p_2021}, the first XLM-RoBERTa-based vision-language encoder, preprinted in the summer of 2020). However, multilingual modeling is critical to the global accessibility of systems based on language models. Joint modeling of language and vision in a multilingual context poses unique challenges that go beyond text-only multilingual modeling.

An important motivation for early research in vision-language modeling was the belief that real-world referential grounding, which is essential for understanding meaning, could not be achieved with text data alone \citep{chrupala-etal-2015-learning,kadar-etal-2018-lessons,hu-etal-2019-looking,tan-bansal-2020-vokenization}. The goal was to use vision to improve language modeling. The main argument was that human language understanding is inseparable from interaction with the physical world, which provides grounding for utterances. Images in vision-language models are seen as the elements providing the necessary grounding through visual representations.

Some researchers take this debate further, arguing that true grounding is unattainable without direct experience with the world and that, without it, learning meaning is fundamentally impossible  \citep{bender-koller-2020-climbing}. Nowadays, there is also substantial counter-evidence against this version of the argument \cite{andreas-2022-language,sogaard2023grounding,gubelmann-2024-pragmatic}, both theoretical and empirical.

Regardless of the theoretical motivations, most practical approaches to vision-language modeling have relied on pre-trained language representations developed without visual data. Similarly, visual representations are often trained using explicit or implicit language-based signals.
However, even image representations are implicitly grounded in language, as they are typically trained on datasets that include language labels or captions (e.g., ImageNet labels, \citealp{deng2009imagenet}, are drawn from the lexical database WordNet, \citealp{miller1995wordnet}). Recent advancements in multimodal large language models (LLMs) have shifted this perspective, with many systems transforming visual representations into the language space and relying solely on language pre-training \citep{eichenberg-etal-2022-magma}.

%The concept of representation alignment has been a recurring theme in multilingual and multimodal research. Many studies have assumed that joint training of language and vision models can lead to improved performance through better-aligned representations \cite{pmlr-v139-radford21a}.

Vision-language models are used in many applications. They include tasks that require reasoning across both textual and visual modalities, such as visual question answering (VQA, \cite{antol2015vqa}) or image captioning. These models have also been important in advancing accessibility technologies, such as image description systems for visually impaired users \citep{merchant2024generating,yang2024viassist}.

\begin{figure}[t]
    \centering
    \includegraphics[width=0.5\linewidth]{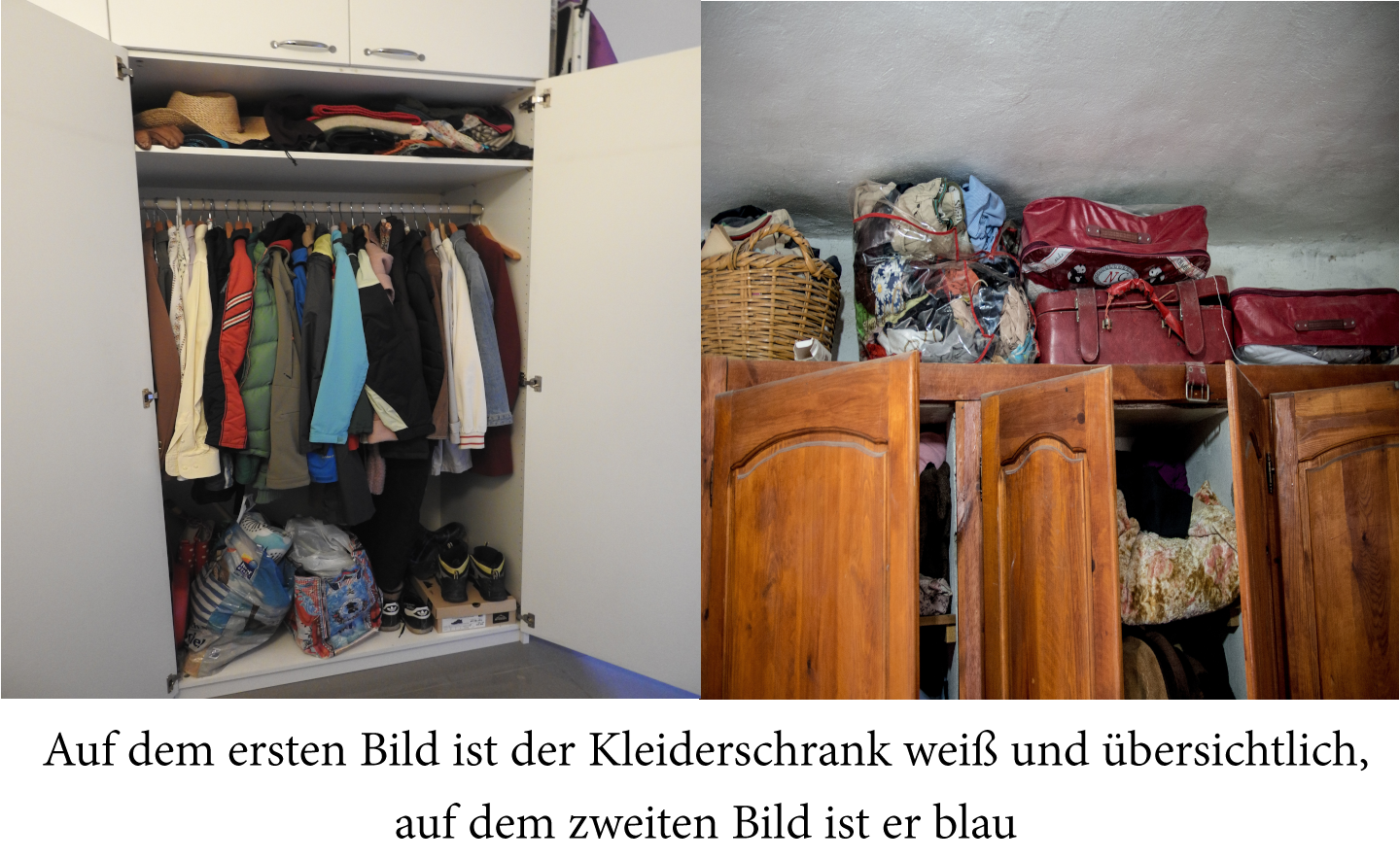}
    \caption{Visual-language reasoning example from M5-VGR set. The original caption in German can be translated as ``In the first picture, the wardrobe is white and clear, in the second picture it is blue''. The answer is ``false''.}
    \label{fig:m5_vgr_sample}
\end{figure}

Most vision-language research focuses on English, and both multilingual models and evaluation benchmarks are derived from English models and datasets.
Maintaining, improving, and even evaluating performance across diverse languages and cultures is thus inherently difficult.
Addressing this challenge is critical for scaling such systems to real-world applications, particularly in global contexts where linguistic and cultural variation must be accounted for \cite{sheng-etal-2021-societal,steed-etal-2022-upstream,karamolegkou-etal-2024-vision}.

In this survey, we critically examine vision-language models from the perspective of two potentially competing principles: \emph{language neutrality} and \emph{cultural awareness}. In the following sections, we define these principles and discuss how they are reflected in training objectives and evaluation benchmarks. We then assess the extent to which current models achieve these goals and the trade-offs involved.

Although we briefly introduce vision-language modeling in general, our focus lies in multilinguality. For a more detailed survey of the English-language vision-language model, we refer readers to existing surveys by \citet{fields_vision_2023} and \citet{akkus_multimodal_2023}.

We structure the article as follows: After discussing language neutrality and cultural awareness in Section~\ref{sec:goals}, we briefly describe the technical prerequisites needed for language vision models in Section~\ref{sec:prerequisites} (language modeling and image representation). In Section~\ref{sec:monolingual}, we provide a basic categorization of monolingual vision-language models. Finally, Section~\ref{sec:models} provides a comprehensive overview of multilingual vision-language models, and Section~\ref{sec:benchmarks} offers a comprehensive overview of multilingual benchmarks.

\section{Conflicting Goals in Multilingual Vision-Language Models}\label{sec:goals}

Developing multilingual vision-language models presents a tension between two competing goals that are rarely mentioned explicitly: \emph{language neutrality} and \emph{cultural awareness}. Language neutrality means representing the same concepts in the same way across languages, ensuring that intersubjective concepts remain consistent regardless of linguistic or cultural context (e.g., a picture of a dog is always a picture of a dog, regardless of the language). Conversely, cultural awareness emphasizes the need to adapt representations and outputs to reflect the cultural and contextual nuances of different languages, in extreme cases requiring distinct interpretations for the same input (in Western culture, dogs are associated with loyalty, whereas in Chinese or Arabic, calling someone a dog might be insulting). This section explores the nature of these conflicting goals and illustrates the challenges they present.

Language neutrality ensures that models produce consistent outputs for objective or universal inputs, regardless of the language used. Examples of such use cases include scientific and factual information (the Pythagorean theorem should be the same in every language, as its truth is invariant to cultural or linguistic differences) and recognizable landmarks (an image of the Eiffel Tower should be recognized as the Eiffel Tower, regardless of the language). In an extreme case, language neutrality would imply removing anything language-specific from the model \citep{libovicky-etal-2020-language} and, in turn, drastically limit the possibility of reaching cultural awareness. Therefore, it is, in practice, replaced by a less strict requirement of cross-lingual alignment \citep{hammerl-etal-2024-understanding} that conceptually relies on representation similarity. It can be a strong alignment that cross-lingually matches concepts to their closest counterparts in the other language, or a weak alignment that only requires the possibility of fine-tuning models for downstream performance across languages, even if the model was fine-tuned in one language.

One of the most common ways to evaluate multilingual and cross-lingual capabilities is zero-shot cross-lingual transfer. In this setup, we fine-tune a multilingual model using training data in one language, and test it on other languages for which the underlying model was pre-trained and were not part of the task-specific fine-tuning. The most common evaluation scheme involves fine-tuning in a high-resource language, typically English, and testing the model on low-resource languages.

In multilingual NLP, some sort of language neutrality is an implicit assumption behind many standard evaluation tasks. For instance, cross-lingual natural language inference (XNLI, \cite{conneau-etal-2018-xnli}), one of the most frequently used cross-lingual evaluation benchmarks, was created by translating the test set of the English MultiNLI \citep{williams-etal-2018-broad} dataset into 14 languages from diverse language families. Thanks to the fact that one of the most famous benchmarks for natural language inference, SNLI \citep{bowman-etal-2015-large}, was created by reusing the sentences from the Flickr30k dataset \citep{young-etal-2014-image}, there is also a vision-language version of this task \citep{xie2019visual}, including its cross-lingual version, which is part of the IGLUE benchmark \citep{bugliarello2022}.
A big advantage of such an approach is that the images remain the same, and the accompanying text is a literal translation of each other. Therefore, the results are directly comparable across languages, and the findings can support claims that a model performs better in one language than in another. However, this might come at the expense of the text not being representative of the languages and cultures in which they are spoken.

The assumption of language neutrality also underlies the problem of multimodal machine translation \citep{specia-etal-2016-shared,elliott-etal-2017-findings,barrault-etal-2018-findings}, which is posed as a literal translation of image captions, focusing on potential ambiguity between languages. This includes both structural language phenomena (e.g., when translating from English into languages with grammatical gender, the image might provide the necessary information) and lexical ambiguities (e.g., the English word ``rock'' can be translated into German and Czech as ``Felsen'' and ``skála'' meaning a rock formation, or as ``Stein'' and ``kámen'' meaning a stone).

The neutrality assumption is crucial for applications that require factual consistency across languages. Additionally, some degree of language neutrality or cross-lingual alignment is required for cross-lingual transfer. However, achieving this goal becomes challenging when concepts or entities lack exact counterparts across languages or are interpreted differently due to cultural factors.

In contrast to language neutrality, \emph{cultural awareness} emphasizes the need to adapt representations and outputs to account for linguistic, cultural, and contextual nuances. It recognizes that language is socially constructed and rooted in the experiences, traditions, and values of its speakers. Being culturally aware often requires distinct outputs for the same input across cultures.

Languages often encode culturally specific information that cannot be directly translated without losing meaning or context. In an American context, a family meal might evoke a Thanksgiving dinner with turkey and pie, while for a user in Japan, it might involve miso soup. Similarly, for an image of a typical home, a German speaker might imagine a European-style house with a gable roof made of clay tiles and windows with potted geraniums and wooden shutters, whereas users in Sub-Saharan Africa might visualize a structure adapted to local climates and materials. Concepts that seem the same are associated with physically and socially distinct objects.

At the same time, developing culturally aware systems comes with risks. One major concern is the potential for harmful stereotyping. While adapting outputs to cultural norms, systems might inadvertently overgeneralize or reinforce biases, portraying cultures in reductive ways. For instance, representing certain regions exclusively through traditional imagery might fail to reflect modern diversity and dynamics.

Conceptualizing culture for model development is not straightforward.
\citet{adilazuarda2024measuringmodelingculturellms} attempts to describe how culture might be operationalized in NLP research. 
%
% Firstly, from a social science perspective, there are multiple definitions (e.g. "The way of life of a collective group of people, [that] distinguishes them from other groups with other cultures"). 
%
They collected 90 articles on text-only datasets evaluating LLMs and found that culture is expressed through demographic (e.g., region, language, ethnicity, religion) and semantic (e.g., emotions, values, food, drink) proxies. There is a noticeable focus on demographic proxies, and we refer readers to their work for further observations.

\begin{figure}[t]
    \centering
    \includegraphics[width=0.5\linewidth]{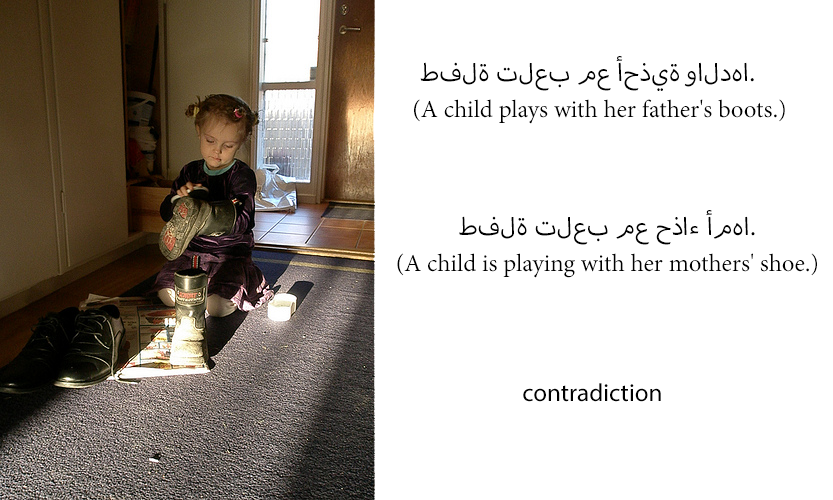}
    \caption{Multilingual natural vision-language inference example from XVNLI set. The first sentence acts like an image caption, while the second one is a different statement, either an ``entailment``, ``contradiction`` or ``neutral`` statement.}
    \label{fig:xvnli_sample}
\end{figure}

As a consequence, evaluating cultural awareness is much more difficult than language neutrality. In tasks such as XNLVI, we have exact semantic equivalents between languages and identical test sets with directly comparable statistics. Having culture-aware but still identical tests is not possible. Benchmarks explicitly focusing on culture-awareness, such as MARVL \citep{liu-etal-2021-visually} for reasoning and XM3600 \citep{thapliyal-etal-2022-crossmodal} for image captioning and retrieval, have images from different regions of the world with local textual counterparts. Even though the benchmark authors aim for comparable test sets, the exact evaluation scores are never directly comparable across languages. This issue becomes more pronounced with generative tasks, where frequently used overlap-based metrics, such as BLEU \citep{papineni-etal-2002-bleu}, chrF \citep{popovic-2015-chrf}, or ROUGE \citep{lin-2004-rouge}, are inherently on different scales across languages. 

Paradoxically, while language neutrality and cultural awareness often appear to be in conflict, they ultimately stem from the same ethical principles: the commitment to equality, fairness, and non-discrimination. Language neutrality ensures that users, regardless of their linguistic background, can access consistent and objective information, aligning with the principle of fairness by treating all languages equally. On the other hand, cultural awareness seeks to provide equitable and meaningful experiences by respecting the contexts of different cultures, avoiding the one-size-fits-all paradigm that has historically marginalized non-Western perspectives.
In the rest of the paper, we use these two perspectives to assess how multilingual vision-language models are trained and evaluated. 

\section{Technical Prerequisities}\label{sec:prerequisites}

Multilingual vision-language models combine techniques from natural language processing and computer vision. To understand how these models work, we first need to cover the key building blocks they use. This section reviews three main components: language modeling for processing text, multilingual modeling for handling multiple languages, and image representation for extracting visual features.

\subsection{Language Modeling}

Language modeling provides the foundation for processing and generating human language by probabilistically predicting word sequences. Modern language modeling employs two primary training paradigms: masked language modeling for encoder architectures and autoregressive language modeling for decoder-based generative models. In both cases, the core training signal derives from predicting missing or subsequent words, serving as a proxy for language understanding. Encoder-decoder models, which are also occasionally used, following the machine translation tradition, typically employ denoising objectives on input text \citep{raffel2023exploringlimitstransferlearning}.

While language modeling has its origins in $n$-gram models \citep{brown-etal-1992-class} and many neural architectures have also been used for language modeling \citep{bengio2003neural,mikolov2011extensions}, the field gained significant momentum with the introduction of the Transformer architecture in 2017 for neural machine translation \citep{vaswani2017attention}. Transformers are neural networks that alternate self-attention layers to capture text structure \citep{marecek-rosa-2019-balustrades,voita-etal-2019-analyzing} and feed-forward layers, interpreted as key-value memories storing linguistic and factual information \citep{geva-etal-2021-transformer}. The first notable pre-trained Transformer language model was BERT \citep{devlin-etal-2019-bert}, which established the paradigm of pre-training on large corpora followed by fine-tuning for downstream tasks such as classification and sequence labeling.
% the field gained, instead of language model field gained

Early generative models, such as the first iteration of GPT \citep{radford2018improving}, initially showed limited utility compared to pre-trained encoders. However, this landscape transformed dramatically through model scaling, as demonstrated by GPT-2 \citep{radford2019language} and GPT-3 \citep{brown2020language}, the discovery of scaling laws \citep{hoffmann2022training}, and advances in instruction fine-tuning \citep{ouyang2022training,rafailov2024direct} that go beyond pre-training with next word prediction.

\subsection{Multilingual Modeling}

Soon after monolingual BERT, multilingual variants appeared. Notable early models include mBERT \citep{devlin-etal-2019-bert} and XLM-RoBERTa \citep{conneau-etal-2020-unsupervised}, which remained state-of-the-art for several years. These massively multilingual encoders feature expanded vocabularies but employ similar training procedures to their monolingual counterparts, with upsampling strategies for low-resource languages. While initially covering approximately 100 languages, subsequent models, such as Glot500 \citep{imanigooghari-etal-2023-glot500}, extend coverage to up to 500 languages.
These models enable cross-lingual transfer, where fine-tuning on one language yields performance gains in other supported languages. \citet{dufter-schutze-2020-identifying} hypothesize that limited parameter constraints force models to derive cross-lingual representations.

While decoder models attempted similar principles through large vocabularies and language balancing, exemplified by XGLM \citep{lin-etal-2022-shot} and BLOOM \citep{scao2022bloom}, they achieved less success compared to contemporary large language models \citep{jiang2023mistral,touvron2023llama2openfoundation,bai2025qwen25vltechnicalreport} that, despite predominantly English training data and fewer parameters, demonstrate strong performance across languages. Training often incorporates parallel data, though performance degrades as digital resources become scarcer for some languages. In extreme cases, models even struggle to distinguish grammatical from ungrammatical sentences \citep{shen2024multiblimp}.

\subsection{Image Representation}

Image representation in vision-language models relies on several architectures that extract visual features. Vision Transformers (ViT;  \citealp{dosovitskiy2020vit}) were invented as a competitive alternative to Convolutional Neural Networks (CNNs). ViT decomposes input images into fixed-size patches treated as sequences of tokens, similar to words in natural language processing, and processes them through transformer encoders. Recent scaling efforts have produced large-scale models such as ViT-22B \citep{dehghani2023scaling}.

Convolutional architectures remain widely used, particularly ResNet variants \citep{he2016deep} were frequently used in vision-language models. EfficientNet \citep{tan2019efficientnet} improves upon that by simultaneously optimizing depth, width, and resolution. These models are typically pre-trained on ImageNet \citep{deng2009imagenet}, a large-scale dataset containing millions of labeled images across thousands of classes.

Some vision-language models do not use features from the entire image, but only representations of objects detected in the image.
For object detection tasks, Faster R-CNN \citep{ren2016fasterrcnnrealtimeobject} integrates CNN backbones with Region Proposal Networks to generate object proposals and classification heads. It processes entire images through the backbone network to produce feature maps, which are subsequently used for proposal generation and object classification.

% \JL{This paragraph is hard to parse.}
% Recently, self-supervised approaches have gained prominence, with models like BEiT \citep{bao2021beit} applying masked image modeling to vision transformers, and BEiT~v2 \citep{peng2022beit} extending this approach with vector-quantized visual tokenizers to enhance semantic representations. These methods are analogous to self-supervised learning in language models like BERT. 
% maybe mention a bit the autoencoders generated visual tokens

\section{Vision-Language Models}\label{sec:monolingual}

Vision-language models combine pre-trained language representations with visual encoders to enable joint processing of textual and visual information. Most vision-language models follow a general architectural pattern: they take a language model, combine it with a pre-trained image representation, and often keep components frozen while training specific fusion layers or cross-modal attention mechanisms. We illustrate this trend in Figure~\ref{fig:timeline}.

Historically, progress in vision-language models typically follows innovations that first appear in language modeling and computer vision.
For instance, BERT \citep{devlin-etal-2019-bert} was introduced in late 2018, with multimodal adaptations like LXMERT \citep{tan-bansal-2019-lxmert} following in mid-2019. Similarly, the scaling successes of models like GPT-3 \citep{brown2020language} in 2020 were followed by large-scale vision-language models like CLIP \citep{pmlr-v139-radford21a} in 2021.

\subsection{Architectures}

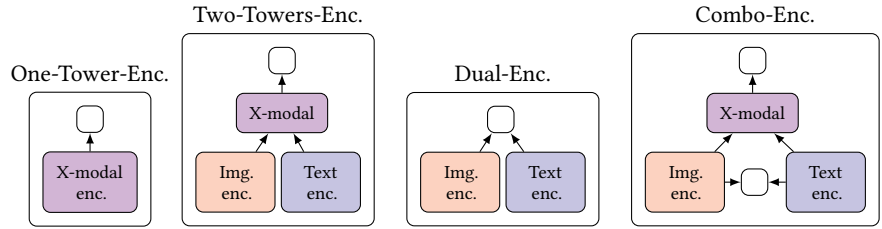
\begin{figure}
\centering{\begin{tikzpicture}[
        mpart/.style={draw, font=\footnotesize, align=center, rounded corners=1mm, inner sep=5pt},
        mbox/.style={draw, rounded corners=1mm, inner sep=5pt},
        databox/.style={draw, cylinder, shape aspect=.5, font=\footnotesize, shape border rotate=90, align=center, cylinder uses custom fill, cylinder body fill=Gray!20, cylinder end fill =Blue!20}]

    \begin{scope}[local bounding box=1tower]
    \node[mpart, text width=12mm, fill=Plum!30] (cm0) {X-modal enc.};
    \node[mpart, above=7pt of cm0] (los0) {};

    \draw[->] (cm0) -- (los0);

    \node[mbox, fit=(cm0) (los0)] {};
    \end{scope}

    \node[above=1pt of 1tower] {One-Tower-Enc.};

    \begin{scope}[
        local bounding box=2towers,
        xshift=55pt
        ]
    \node[mpart, text width=7mm, fill=Red!20] (img1) {Img. enc.};
    \node[mpart, right=2pt of img1, text width=7mm, fill=Blue!20] (txt1) {Text enc.};
    \node[mpart, above right=7pt and -15pt of img1, fill=Plum!30] (cm1) {X-modal};
    \node[mpart, above=7pt of cm1] (los1) {};

    \draw[->] (img1) -- (cm1);
    \draw[->] (txt1) -- (cm1);
    \draw[->] (cm1) -- (los1);

    \node[mbox, fit=(img1) (txt1) (cm1) (los1)] {};
    \end{scope}

    \node[above=1pt of 2towers] {Two-Towers-Enc.};

    \begin{scope}[
        local bounding box=dual_enc,
        xshift=140pt
        ]
    \node[mpart, text width=7mm, fill=Red!20] (img2) {Img. enc.};
    \node[mpart, right=2pt of img2, text width=7mm, fill=Blue!20] (txt2) {Text enc.};
    \node[mpart, above right=7pt and -5pt of img2] (los2) {};

    \draw[->] (img2) -- (los2);
    \draw[->] (txt2) -- (los2);

    \node[mbox, fit=(img2) (txt2) (los2)] {};
    \end{scope}

    \node[above=1pt of dual_enc] {Dual-Enc.};

    \begin{scope}[
        local bounding box=combo_enc,
        xshift=225pt
        ]
    \node[mpart, text width=7mm, fill=Red!20] (img3) {Img. enc.};
    \node[mpart, right=23pt of img3, text width=7mm, fill=Blue!20] (txt3) {Text enc.};
    \node[mpart, above right=7pt and -7pt of img3, fill=Plum!30] (cm3) {X-modal};
    \node[mpart, above=7pt of cm3] (los3) {};
    \node[mpart, right=6pt of img3] (los4) {};

    \draw[->] (img3) -- (cm3);
    \draw[->] (txt3) -- (cm3);
    \draw[->] (cm3) -- (los3);
    \draw[->] (img3) -- (los4);
    \draw[->] (txt3) -- (los4);

    \node[mbox, fit=(img3) (txt3) (cm3) (los3) (los4)] {};
    \end{scope}
    
    \node[above=1pt of combo_enc] {Combo-Enc.};

\end{tikzpicture}}
\caption{Overview of the VL encoders architecture. The white empty box represents a nontrainable layer that computes the loss function.}
\label{fig:1}
\end{figure}

\begin{figure}
\centering{\input{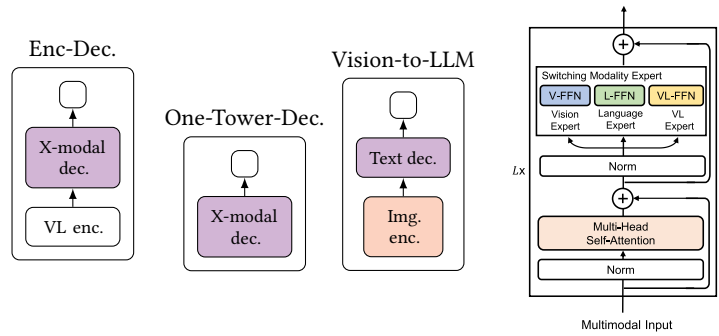}}
\caption{Overview of the VL decoder's architecture and the transformer layer for a mixture of modality experts, picture extracted from VLMo. The white empty box represents a nontrainable layer that computes the loss function. The \textit{VL enc.} can be any encoder from Figure~\ref{fig:1}.}
\label{fig:2}
\end{figure}

The field has developed several distinct architectural approaches for fusing visual and textual information. Following the categorization by \citet{fields_vision_2023} for English-only models, we identify multiple architecture types:

\paragraph{One-Tower Encoders.} These architectures combine embeddings from both images and text into a single input stream processed by a shared encoder, treating them as a single sequence so that the model shares parameters for text and image contextualization. This approach is used in models like UNITER \citep{vedaldi_uniter_2020}, UC2 \citep{zhou_uc2_2021}, and M3P \citep{ni_m3p_2021}.

% Masked cross-modal modeling requires comparable tokens and patches.

\paragraph{Two-Tower Encoders.} These models employ separate encoders for images and text, with their intermediate outputs passed to a cross-modal encoder. Models like LXMERT \citep{tan-bansal-2019-lxmert}, UNITER variations \citep{kwon_masked_2023}, and BridgeTower \citep{xu_bridgetower_2023} use this design. Sequences of text embeddings and image region representations are processed by independent Transformer encoders with a distinct set of parameters. The cross-modal encoder applies attention mechanisms where one modality generates queries while the other provides keys and values, facilitating nuanced interactions between visual and textual representations. Even if this is more parameter-efficient, it limits the interaction between images and text in the first layers.

\paragraph{Dual Encoders.} These architectures process each modality separately and combine them only at the final stage via similarity measures or joint embeddings. CLIP \citep{pmlr-v139-radford21a}, ALIGN \citep{pmlr-v139-jia21b}, and MURAL \citep{jain-etal-2021-mural-multimodal} are examples of this approach. They are particularly effective for retrieval tasks because they allow for the pre-computation of the unimodal representation. However, in the reasoning tasks, these models perform worse because they do not allow interaction between the text and visual inputs until the last layer.

\paragraph{Encoder-Decoders.} These models go beyond representation learning to include generative capabilities. Models like BLIP \citep{li_blip_2022}, SimVLM \citep{wang2022simvlmsimplevisuallanguage}, and MAPL \citep{manas-etal-2023-mapl} incorporate decoders for tasks requiring text generation, such as image captioning or open-ended visual question answering.

\vspace{.5\baselineskip}

\noindent Beyond these established categories, we identify additional architectural types:

\paragraph{Combo Encoders.} These models represent a variant of two-tower architectures where fusion is performed not only within the cross-modal encoder but also between the unimodal encoder representations using an additional loss, as in dual encoders. ALBEF \citep{li_align_2021}, X2-VLM \citep{zeng_x2-vlm_2023}, and related models \citep{zeng-etal-2023-cross} use multiple fusion points with different loss functions to improve multimodal alignment.

\paragraph{Mixture-of-Modality-Experts.} This class of models uses modality-specific modules within transformer architectures. VLMo \citep{bao_vlmo_2022} and BEiT \citep{wang_image_2022} replace standard feed-forward layers with three specialized layers for vision, text, and cross-modal representations, along with task-dependent switching mechanisms. This approach addresses the parameter inefficiencies of two-tower encoders while maintaining the possibility of cross-modal interaction.

\paragraph{Decoder Models.} These models extend classical generative large language models by integrating vision components. Models like MAGMA \citep{eichenberg-etal-2022-magma}, mBLIP \citep{geigle-etal-2024-mblip}, and others \citep{liu2024improvedbaselinesvisualinstruction, manas-etal-2023-mapl} map visual features into the language model space, enabling multimodal input processing within generative frameworks. Because all parameters are shared for both text and vision modalities, they can be considered generative analogs to one-tower encoders.

\vspace{.5\baselineskip}

\noindent In addition to categorizing models by how they combine modalities, vision-language models can also be classified by how they represent and process visual information. There are three main approaches:
\begin{itemize}

\item \emph{Multimodal Single Streams.} These models concatenate visual and textual features into a unified input sequence that is processed by a single transformer architecture. Visual features are typically encoded and treated as special tokens that can be intermixed with text tokens. MAGMA \citep{eichenberg-etal-2022-magma} exemplifies this approach by creating joint representations directly from the unified multimodal input stream.

\item \emph{Vision-to-LLM Mapping.} This category includes models that map visual representations into the language model's embedding space through two distinct strategies.

\begin{itemize}

\item \emph{Continuous representations.} Visual features are mapped into a continuous numerical space that aligns with language model embeddings, allowing seamless integration without discretization. Examples include MAPL \citep{manas-etal-2023-mapl} and mBLIP \citep{geigle-etal-2024-mblip}.

\item \emph{Discrete tokens.} Visual features are quantized into discrete units that mimic the structure of text tokens, enabling direct processing by language models designed for discrete inputs. This approach is adopted by models such as SEED \citep{ge2023makingllamadrawseed}, LaVIT \citep{zhan-etal-2024-anygpt}, and AnyGPT \citep{jin2024unified}.

\end{itemize}

\item \emph{Hybrid Fusion.} A combination of the two mentioned approaches, within the submodules. For instance, one module might process concatenated multimodal streams, while another handles discrete visual tokens, with outputs integrated for final predictions \citep{diao2023writepaintgenerativevisionlanguage}.

\end{itemize}

Among the most influential models, CLIP \citep{pmlr-v139-radford21a} introduced a dual-encoder architecture trained with contrastive learning on 400 million image-text pairs from the web. The model uses separate encoders for images (typically Vision Transformers or ResNets) and text (GPT-like transformers), learning to maximize cosine similarity between matching image-text pairs while minimizing similarity between non-matching pairs. UNITER \citep{vedaldi_uniter_2020} employs a one-tower approach with four pre-training tasks: masked language modeling, masked region modeling, image-text matching, and word-region alignment using optimal transport. ALBEF \citep{li_align_2021} combines the benefits of contrastive learning and cross-modal fusion by first aligning image and text representations through contrastive loss before fusing them via cross-modal attention, additionally introducing momentum distillation to handle noisy web data.

\subsection{Data and Pre-training}

Vision-language models typically employ several pre-training objectives adapted from both computer vision and natural language processing. The most important training tasks include:

\paragraph{Masked Language Modeling (MLM).} This approach directly adapts BERT's training objective to multimodal settings. The model receives sentences or multimodal inputs with randomly masked text tokens and must predict the masked tokens using both textual context and visual information. Even if the training starts with a pre-trained text encoder, MLM is used for continuous training. Most encoder models, such as UNITER \citep{vedaldi_uniter_2020}, ALBEF \citep{li_align_2021}, and VisualBERT \citep{li2019visualbert}, use MLM.

\paragraph{Masked Region Modeling (MRM).} This serves as the visual counterpart to MLM. The model receives images with specific regions covered by binary masks and must reconstruct the masked regions. Reconstruction targets can range from raw pixels to higher-level visual features such as object categories or mean color values. UNITER \citep{vedaldi_uniter_2020} and VL-BERT \citep{su2019vl} incorporate MRM variants in their training.

\paragraph{Image-Text Matching (ITM).} This task requires the model to determine whether a given image and caption correspond to each other. This binary classification task helps the model learn global correspondences between visual and textual content. UNITER \citep{vedaldi_uniter_2020}, ALBEF \citep{li_align_2021}, and BLIP \citep{li_blip_2022} utilize ITM as part of their multi-task training.

\paragraph{Contrastive Learning (CL).} This approach learns to align positive image-text pairs while separating negative pairs within a batch. Given a batch of image-text pairs, the model maximizes similarity between semantically related pairs while minimizing similarity between unrelated pairs. This approach, popularized by CLIP \citep{pmlr-v139-radford21a}, has proven highly effective for learning aligned multimodal representations. Other models adopting contrastive learning include ALIGN \citep{pmlr-v139-jia21b} and ALBEF \citep{li_align_2021}. As an alternative to the softmax normalization function, \citet{zhai2023sigmoidlosslanguageimage} proposed sigmoid normalization in SigLIP as a more efficient approach capable of handling larger batches of image-text representation pairs.

Additional specialized objectives include optimal transport-based word-region alignment \citep{vedaldi_uniter_2020}, momentum distillation for handling noisy data \citep{li_align_2021}, and various generative objectives for models with decoder components \citep{li_blip_2022, wang2022simvlmsimplevisuallanguage}. The choice of pre-training objectives significantly influences model capabilities and downstream performance across different task categories.

\subsection{Standard Evaluation Tasks}

Vision-language models are evaluated across three main categories of downstream tasks that assess different aspects of multimodal understanding. In this overview, we focus on tasks that are also relevant in the multilingual context.

\paragraph{Multimodal Reasoning.} These tasks assess the model's capability to grasp relationships between elements in images and text at logical and semantic levels. The tasks are typically formulated as classifications, which allows for straightforward evaluation using accuracy and the F1-score.

\begin{itemize}

\item Visual Question Answering \citep{hudson_2019_CVPR, antol2015vqa} --- using an image and a question, the model provides a one-word answer, either from a predefined list given during pre-training or from a list of options provided in the textual input immediately following the question. This is the scenario for a multilabel classification task. At the same time, there is open-ended VQA, where the model has to generate an answer that is then fed into a machine translation metric, such as BLEU \citep{papineni-etal-2002-bleu}, chrF \citep{popovic-2015-chrf}, and LLM-as-a-judge \citep{zheng2023judge}. These metrics are less precise since matching the meaning of two distinct sentences relies either on the overlap of characters or on the "LLM point of view".

\item Natural Language Vision Inference \citep{xie2019visual} --- considering a contextual image and 2 captions, the model outputs the relationship between the 2 sentences as either logical entailment, contradiction, or neutrality. This is an adaptation of a popular NLP task in Natural Language Inference \citep{bowman-etal-2015-large}.

\item Natural Vision-Language Reasoning \citep{suhr-etal-2019-corpus} --- using two images and one caption that describes one visual aspect, the model answers either True or False.

\item Visual Commonsense Reasoning \citep{zellers_2019_CVPR} --- considering not only an image but also a list of regions in the image, the model has to answer multiple-choice questions.

\item Image Captioning \citep{karpathy_2015_CVPR, chen2015microsoftcococaptionsdata} --- can only be tested with encoder-decoder and decoder-only models, computing the image representation and translating it into descriptive text. Unlike previous tasks, image captioning is an open-ended generation task, which makes automatic evaluation more difficult. The usual evaluation metrics include BLEU \citep{papineni-etal-2002-bleu}, chrF \citep{popovic-2015-chrf}, or CiDER score \citep{vedantam2015cider}.

\end{itemize}

\begin{figure}
\centering
\scalebox{0.80}{\rotatebox{270}{\begin{minipage}{218mm}\centering \input{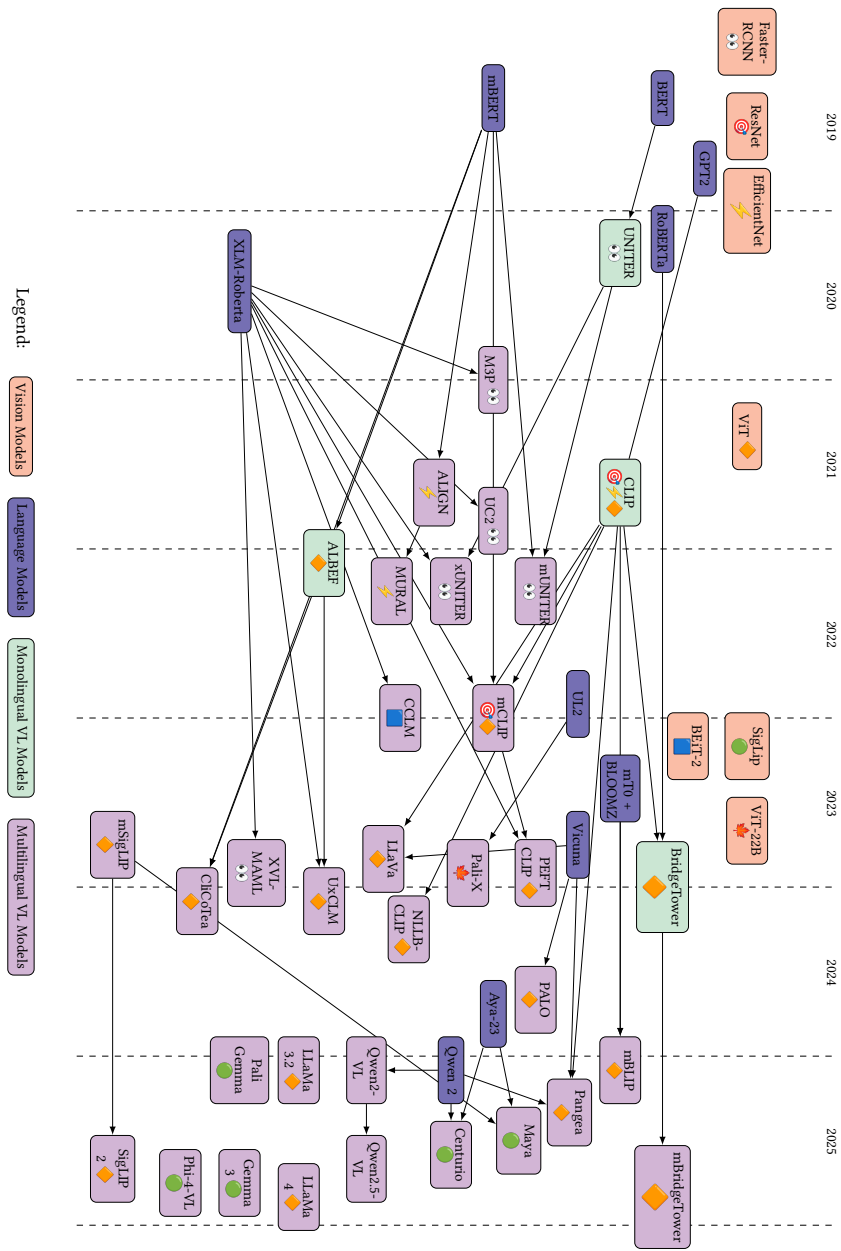}\end{minipage}}}

\caption{Timeline and dependencies of models discussed in this survey. Models are ordered by release date. The arrows in the diagram indicate that the model was developed by reusing or finetuning parts of a previous model. For clarity, we indicate the used vision encoder using emojis.}
\label{fig:timeline}
\end{figure}

\paragraph{Cross-modal Similarity.} Here, the focus is on understanding the correspondence and similarity relationships between images and texts.
It is typically evaluated using Image-Text Retrieval \citep{young-etal-2014-image, chen2015microsoftcococaptionsdata}. In this setup, we compute image and text representations, retrieve images from text prompts, and vice versa. Retrieval is typically evaluated using top-1 accuracy.

\paragraph{Computer Vision Tasks.} Vision-language models are often evaluated on tasks that are traditionally considered to belong to computer vision. These tasks emphasize visual details, often requiring precise localization or attribute recognition.

\begin{itemize}

\item Image Classification, Object Detection, Semantic Segmentation --- pure computer vision tasks that evaluate visual understanding capabilities.

\item Visual Grounding \citep{kazemzadeh-etal-2014-referitgame} --- using an image and a query to search for a visual concept, the model must predict its bounding box.

\item Referring Expression Comprehension \citep{kazemzadeh-etal-2014-referitgame} --- retrieval tasks applied to proposed regions and images; performance is measured based on both the recognized region and the intersection over union of the bounding box.

\item Open Vocabulary Attribute Detection \citep{bravo_2023_CVPR} --- object detection that requires identifying additional attributes such as color, texture, and other fine-grained optical properties.

\end{itemize}

Additionally, specialized benchmarks evaluate downstream tasks across different domains that go beyond the semantic understanding of scene images and focus on the visual understanding of documents. Examples of such tasks are SEED-Bench for VQA \citep{li2023seedbenchbenchmarkingmultimodalllms}, which covers multiple domain dimensions; MathVista \citep{lu2024mathvistaevaluatingmathematicalreasoning}, which specializes in mathematical reasoning; and ChartQA \citep{masry-etal-2022-chartqa}, which focuses on charts and diagrams.

There is a recent trend in evaluating LLMs with mathematical, coding reasoning, and other scientifically grounded tasks. However, we consider that reasoning in natural language is still a challenging task, especially when grounded in a cultural context.

\section{Multilingual Vision-Language Models}\label{sec:models}

We report the most important recent models in Table~\ref{tab_mm_ml_models} and in the timeline diagram in Figure~\ref{fig:timeline}. We consider only models with open weights. The field has witnessed rapid evolution from early encoder-only models, primarily for classification tasks in 2020, to current decoder-only architectures, with a shift toward larger parameter counts and generative tasks.

In Table~\ref{tab_mm_ml_models}, we show basic descriptive statistics of the models. Models marked with the \textit{ad} superscript were obtained through cross-lingual transfer or adaptation rather than end-to-end multilingual pre-training.
It is not straightforward to determine how many languages the models support. Not only do models perform differently across languages, but distinct languages can be used at different stages of model development.
The underlying multilingual text models implicitly support a number of languages due to language-only pre-training, typically ranging from dozens to over 200 languages for some recent models. We report this as the number of implicitly supported languages. The number of explicitly supported languages refers to those for which the model has been specifically pre-trained or adapted with multimodal data. Finally, the number of evaluation languages indicates the scope of empirical assessment in published benchmarks. Most model weights are released on GitHub, and many are also available on the Hugging Face Hub. For those also available on Hugging Face, we display download counts (last updated 29.01.2026) to reflect public adoption. Moreover, we manually add the number of article citations from Google Scholar and the average accuracy on MARVL.

\subsection{General Trends and Observations}

\begin{table*}
\caption{Overview of the recent multilingual multimodal models with their architecture (Enc stands for Encoder, Dec stands for Dec), number of languages (implicitly in the language-model training, or explicitly in multimodal training and in the reported model evaluation), number of citations in Google Scholar, their demand on HuggingFace, and accuracy on the MARVL dataset if reported.}
\centering
% \begin{tabular}{l c c@{\hskip-10pt}c@{\hskip-12pt} c@{\hskip-12pt} c@{\hskip-12pt} c@{\hskip-2pt}c@{\hskip-12pt}c}
\setlength{\tabcolsep}{3.6pt}
\begin{tabular}{l cccccccc}
\toprule
Model & Architecture & \# params & \focuscell{\# implict \\ langs} & \focuscell{\# explicit \\ langs} & \focuscell{\# eval \\ langs} & \# cite & \focuscell{\# HF \\ downloads} & \focuscell{MARVL \\ avg. acc.} \\ \midrule

UC2 		 & 1Tower-Enc & 282M & 100 & 6   & 6 & 124 & --- & 57.3 \\
M3P 		 & " 			  				   & 377M & 100 & 100 & 6 & 146 & --- & 56.0 \\
mUNITER 	 &  "			  				   & 137M & 104 & 104 & 6 & 234 & --- & 53.7 \\
xUNITER 	 & " 			  				   & 152M & 100 & 104 & 6 & 234 & --- & 54.6 \\
\midrule

$\text{mBridgeTower}^{ad}$ & 2Tower-Enc 				   & 330M & 100 & 44 & 46 & 0 & --- & 66.7 \\
\midrule

CCLM 		            & Combo-Enc                          & 420M & 100 &  21 & 21 & 43 & --- & 67.2 \\
*UxCLM*  	            & " 								 & 377M & 100 & 100 & 21 & 28 & --- & 62.1 \\
$\text{CliCoTea}^{ad}$ 	& "								     & 210M & 104 &  1* & 12 & 6 & --- & 68.1 \\
SigLIP-2 	            &  "								 & 86M - 1B & 109 & 109 & 36 & 427 & 5M & --- \\
\midrule

ALIGN 		                & Dual Enc                              & 300M & 104 & 100+ &   4 & 5904 & ---  & --- \\
MURAL 		                & "								        & 300M & 104 & 124  & 108 & 100   & ---  & --- \\
mSigLIP                     & "                                     & 400M & 101 & 109  &  36 & 2516 & 3M   & --- \\
$\text{XVL-MAML}^{ad}$      &  "                                    & 282M & 100 & 14   &  14 & 4    & ---  & 59.7 \\
$\text{mCLIP}^{ad}$         &  "	    						    & 178M & 104 & 68   &  12 & 148  & 1M & --- \\
$\text{PEFT-CLIP}^{ad}$ 	& " 		    				  	    & 178M & 100 & 12   &  12 & 6    & ---  & --- \\
$\text{NLLB-CLIP}^{ad}$	    &	"							        & 507M & 200 & 200  & 200 & 22   & 20K  & --- \\
\midrule

mBLIP mT0-XL & Enc-Dec                          & 4.9B & 101 & 96 & 39 & 44 & 24K & 75.1 \\
Pali-X 		 &	"								& 24.5B & 1 & 100 & 36 & 283 & --- & ---\\
T5Gemma 2    &  "                               & 270M - 4B & 140 & 140 & 101 & 0 & 25K & --- \\
\midrule

mBLIP BLOOMZ-7B             & Dec-Only                              & 8.3B & 46 & 96 & 39 & 44 & 4K & 73.9 \\
LLaVa 1.5            		&  "								   & 7B & 3 & 3 & 4 & 4383 & 3M & --- \\
LLaMA 3.2            		&  "								   & 11B & 8 & 1 & 1 & 8409 & 17M & --- \\
Maya 						&  	"							         & 8B & 23 & 8 & 8 & 1 & 2K & --- \\
PaliGemma-3B 				&  	"							       & 2.9B & 1 & 109 & 37 & 522 & 5M & 80.6 \\
Centurio					&  	"							       & 3.8B & 11 & 100 & 100 & 8 & 495 & 81.7 \\
Palo 						&  	"							       & 13B & 3 & 10 & 10 & 28 & 373 & --- \\
Pangea  					&  "								   & 7B & 3 & 39 & 46 & 60 & 166K & 79.0 \\
Qwen-2 VL 					&  "								   & 2B - 72B & 7 & 9 & 9 & 2440 & 268K & --- \\
Qwen-2.5 VL                 &  "                                    & 3B - 72B & 7 & 10 & 1 & 3462 & 43M & --- \\
Phi-4-Multimodal 			&  "								   & 3.8B & 1 & 9 & 10 & 219 & 5M & --- \\
GEMMA 3 					&  "								   & 1B - 27B & 109 & 140 & 922 & 62 & 9M & --- \\
LLaMa 4 					&  	MoME                               & 109B - 2T & 8 & 200 & 14 & --- & 4M & --- \\
Qwen-3 VL                   & Dec + MoME                            & 0.6B - 235B & 29 & 119	& 119 & 0 & 7M & --- \\

\bottomrule
\end{tabular}
\label{tab_mm_ml_models}   
\end{table*}

Our survey covers 33 multilingual vision-language models spanning from 2020 to 2025, revealing several evolutionary patterns in the field. Architecturally, we observe a clear progression from encoder-based models (16 models) to decoder-only architectures (14 models), with fewer encoder-decoder (3 models). This shift mirrors broader trends in NLP.

Parameter scales have increased dramatically over this period. Early encoder models typically contained 100-–400 million parameters, while recent decoder-only models span from 1 billion to 2 trillion parameters in the case of the mixture-of-expert LLaMA 4.

Language coverage patterns reveal interesting disparities between implicit and explicit multilingual support. While many models implicitly support 100+ languages through their underlying language encoders (particularly those based on XLM-RoBERTa or mBERT), explicit multilingual training data often covers far fewer languages. For instance, mUNITER and xUNITER \citep{liu-etal-2021-visually} rely purely on the multilingual capabilities of pre-trained representations without using any multilingual multimodal data, whereas models like MURAL \citep{jain-etal-2021-mural-multimodal} and mSigLIP \citep{zhai2023sigmoidlosslanguageimage} incorporate large-scale multilingual collections spanning more than 100 languages.

Training data strategies include many approaches. Some models use minimal multilingual adaptation, e.g., through synthetic code-switching \citep{ni_m3p_2021} or small-scale parallel datasets \citep{zeng-etal-2023-cross} (CCLM with 19M parallel sentences), while others leverage massive multilingual collections \citep{pmlr-v139-jia21b} (ALIGN with 1.8B image-text pairs, mSigLIP \citep{zhai2023sigmoidlosslanguageimage} with 30B pairs).
Commercial open-weight models lack transparency in this regard, as technical reports focus primarily on architectural innovations and provide limited details on the composition of training data and language distribution.

Evaluation practices typically assess performance in fewer languages than those supported during training. Most benchmarks evaluate 4--46 languages, significantly fewer than the 100+ languages that many models theoretically support. This evaluation gap highlights the relative scarcity of multilingual vision-language benchmarks and the field's continued bias toward high-resource languages. We discuss the evaluation in more detail in Section~\ref{sec:benchmarks}.

Typically, the development of a multilingual vision-language model starts with a multilingual language-only model, which is then augmented with multimodal capabilities through continued pre-training with pre-trained vision representations. Some models take the opposite approach: they start with an English-only vision-language model and extend to multiple languages by replacing the text encoder. Models like mCLIP \citep{carlsson-etal-2022-cross}, PEFT-CLIP \citep{zhang-etal-2023-parameter}, and CliCoTea \citep{karoui-etal-2023-stop} demonstrate that existing English vision-language models can be successfully adapted through knowledge distillation or parameter-efficient fine-tuning with competitive performance.

As we discussed in Section~\ref{sec:goals}, there are two conflicting requirements for multilingual vision-language models: language neutrality and cultural awareness. Language neutrality can be easily incorporated into training objectives to enforce consistent representations across languages; for instance, models that employ contrastive learning on translated captions, as we often see in encoder models (e.g., NLLB-CLIP \citep{visheratin2023}, CliCoTea \citep{karoui-etal-2023-stop}). However, there is no work on culturally aware training objectives, so it is not clear how culture can be integrated, apart from including authentic images from different cultural contexts and authentic descriptions in the local languages. Such is the case for ALIGN \citep{pmlr-v139-jia21b}, mBLIP \citep{geigle-etal-2024-mblip}, Centurio \citep{geigle2025centuriodriversmultilingualability}. For commercial models, this information is often undisclosed.

\subsection{Detailed Model Overview}

In this section, we provide a brief description of each of the models from Table~\ref{tab_mm_ml_models} and Figure~\ref{fig:timeline}. The models are discussed in chronological order of their introduction. For quicker orientation, we graphically display the model architecture (one-tower encoder \badgeOneTower, two-tower encoder \badgeTwoTower, combo encoder \badgeComboEncoder, dual encoder \badgeDualEncoder, encoder-decoder \badgeEncoderDecoder, and decoder-only \badgeDecoder), the number of parameters \badgeParams{100M}, and the month of release \badgeDate{01/2026}.

Readers who are not interested in the technical details of the models can skip to Section~\ref{sec:benchmarks}, which discusses evaluation benchmarks.

\paragraph{M3P.} \badgeOneTower{}~\badgeParams{377M}~\badgeDate{06/2020}\quad \citet{ni_m3p_2021} introduced one of the earliest attempts to create a unified multilingual VL encoder using multitask pre-training. The model uses a one-tower architecture that processes three distinct types of input: multilingual text, synthetic code-switched text with images, and monolingual texts with images. Visual features are extracted using Faster R-CNN \citep{ren2016fasterrcnnrealtimeobject} with region proposals. M3P used synthetic code switching, randomly replacing tokens in sentences with their translations; an innovative method that was not used in later models.

%A key innovation in M3P is Multimodal Code-switched Training (MCT), which randomly replaces tokens in sentences with their translations in different languages to encourage fine-grained alignment between images and non-English languages. This code-switching strategy operates exclusively on the textual modality and aims to map objects across different modalities and texts in various languages into a common semantic space.

\paragraph{UC2.} \badgeOneTower{}~\badgeParams{550M}~\badgeDate{04/2021}\quad \citet{zhou_uc2_2021} builds a multilingual VL encoder by using authentic multilingual sentences. Their approach involves the machine translation of the Conceptual Caption dataset \citep{sharma2018conceptual} into six languages and the use of additional pre-training objectives. The first objective, Masked Region-to-Token Modeling, requires the model to classify masked image regions using pseudo-labels generated by external object detectors. The second innovation, Early Adaptation, introduces a linear projection layer that maps visual region representations closer to the token embedding space by minimizing cosine similarity between visual features and corresponding text token embeddings from the same object detector labels. As M3P, it uses XLM-R to initialize the language encoder.

\paragraph{mUNITER and xUNITER.} \badgeTwoTower{}~\badgeParams{350M}~\badgeDate{09/2021}\quad Concurrent with the release of the MARVL dataset, \citet{liu-etal-2021-visually} develops two multilingual extensions of UNITER \citep{vedaldi_uniter_2020}. These models maintain the original UNITER architecture and replace the text encoder: mUNITER substitutes BERT with multilingual BERT \citep{devlin-etal-2019-bert}, while xUNITER uses XLM-R \citep{conneau-etal-2020-unsupervised}. Unlike previous models, it does not use multilingual training data and relies solely on XLM-R's multilingual capabilities in a zero-shot setup.

\paragraph{ALIGN.} \badgeDualEncoder{}~\badgeParams{600M}~\badgeDate{02/2021}\quad \citet{pmlr-v139-jia21b} developed ALIGN as a dual-encoder model trained using contrastive learning objectives similar to CLIP \citep{pmlr-v139-radford21a}. The model was trained on a single large dataset of 1.8 billion image-text pairs spanning more than 100 languages. It uses EfficientNet-based visual representations.

\paragraph{MURAL.} \badgeDualEncoder{}~\badgeParams{600M}~\badgeDate{09/2021}\quad \citet{jain-etal-2021-mural-multimodal} made an extension of ALIGN with 6 billion parallel text pairs across different languages. These parallel text pairs were generated using machine translation, and the model employs the same dual-encoder architecture and pre-trained weights as ALIGN. The additional cross-lingual contrastive objective aims to improve the alignment between languages in the shared embedding space.

\paragraph{CCLM.} \badgeComboEncoder{}~\badgeParams{560M}~\badgeDate{06/2022}\quad \citet{zeng-etal-2023-cross} uses the X$^2$-VLM \citep{zeng_x2-vlm_2023} architecture with XLM-R \citep{conneau-etal-2020-unsupervised} for text processing and a dedicated cross-modal encoder. The model accepts either image-text or bilingual text pairs as input, with unimodal representations feeding into contrastive loss functions. The cross-modal encoder implements a two-tower logic with cross-attention mechanisms between modalities. X$^2$-VLM was pre-trained on 1B image-text pairs; the multilingual training uses only 19M parallel sentences.

\paragraph{mCLIP.} \badgeDualEncoder{}~\badgeParams{420M}~\badgeDate{6/2022}\quad \citet{carlsson-etal-2022-cross} presented a computationally inexpensive method of adapting existing English CLIP models rather than pre-training from scratch. Their approach involves replacing CLIP's English text encoder with multilingual alternatives such as multilingual BERT, XLM-R, or Swedish BERT. During adaptation, the multilingual model (student) learns to replicate the representation of the original English encoder (teacher). 

\paragraph{PEFT-CLIP.} \badgeDualEncoder{}~\badgeParams{420M}~\badgeDate{05/2023}\quad \citet{zhang-etal-2023-parameter} identified performance disparities across languages in mCLIP due to uneven language distribution during pre-training. They addressed this by using machine translation to generate additional parallel sentences and employing parameter-efficient fine-tuning techniques. Their training objective combines the original cross-lingual loss with a mean-squared-error term between the translated and original sentence representations. Despite these innovations, improvements on retrieval datasets are rather small compared to mCLIP.

\paragraph{NLLB-CLIP.} \badgeDualEncoder{}~\badgeParams{420M}~\badgeDate{09/2023}\quad \citet{visheratin2023} used the NLLB translation model \citep{nllbteam2022languageleftbehindscaling} to extend CLIP to the 200 languages covered by NLLB. They machine-translated the retrieval datasets XTD \citep{aggarwal2020} and Flickr30k \citep{young-etal-2014-image}, replaced the text encoder with LASER NLLB, and fine-tuned the entire model on the expanded multilingual datasets. In this way, they reached substantial improvements for the low-resource languages of XM3600. However, on XTD, they did not outperform the state-of-the art models.
% \JL{Find and criticize how they evaluate.}
% the drawback is that they rely on MT a lot...
% on XTD: "NLLB-CLIP did not outperform state-of-the-art models, the large model is not very far behind - 90.1% vs. 93.7% on average"

\paragraph{UxCLM.} \badgeComboEncoder{}~\badgeParams{550M}~\badgeDate{07/2023}\quad \citet{li-etal-2023-unifying} proposed a weakly supervised multilingual VL encoder trained on 4M image-text pairs, 19M parallel translation pairs, and 800M unlabeled sentences. The model combines contrastive learning on multilingual and multimodal data with masked language modeling applied across all input types (multilingual sentences, parallel pairs, and image-text pairs). The approach showed substantial improvements over previous multilingual VL encoders on the IGLUE benchmark \citep{bugliarello2022}.

\paragraph{XVL-MAML.} \badgeTwoTower{}~\badgeParams{350M-550M}~\badgeDate{05/2023}\quad \citet{hu_meta-learning_2023} adapted the X-MAML meta-learning algorithm \citep{nooralahzadeh-etal-2020-zero} for multimodal settings in cross-lingual VL transfer. The approach trains models (xUNITER and UC2) by splitting each language's dataset into support and query sets, updating parameters on the support set, and then evaluating on the query set. This process iterates across multiple language tasks to learn transferable initial parameters.

\paragraph{CliCoTea.} \badgeComboEncoder{}~\badgeParams{450M}~\badgeDate{07/2023}\quad \citet{karoui-etal-2023-stop} proposed a cross-lingual transfer framework that adapts English VL encoders to other languages. The approach replaces ALBEF's \citep{li_align_2021} English BERT encoder with multilingual BERT and trains on 200,000 machine-translated parallel samples per target language from IGLUE \citep{bugliarello2022}. The multilingual model learns to minimize the mean squared error between its token embeddings and those of the English model, using token-level alignments computed via Awesome Align \citep{dou-neubig-2021-word}.

\paragraph{mBridgeTower.} \badgeTwoTower{}~\badgeParams{866M}~\badgeDate{04/2025}\quad \citet{manea2025BridgeTower} extended BridgeTower \citep{xu_bridgetower_2023} to multilingual settings by combining training objectives from CliCoTea \citep{karoui-etal-2023-stop} and mCLIP \citep{carlsson-etal-2022-cross}. The model is trained on various parallel datasets, with ablations showing that translated task-specific data provides the largest gains. Results on reasoning datasets (MARVL, M5-VGR) approach but do not exceed state-of-the-art performance.

\paragraph{mSigLIP.} \badgeDualEncoder{}~\badgeParams{400M}~\badgeDate{03/2023}\quad \citet{zhai2023sigmoidlosslanguageimage} replaced CLIP's softmax-based contrastive learning with element-wise sigmoid operations, improving efficiency for batch sizes below 16,000. The model was trained on 30 billion image-text pairs from WebLI across 109 languages. To handle large multilingual vocabularies, mSigLIP factors the token embeddings into two smaller matrices.

\paragraph{SigLIP 2.} \badgeDualEncoder{} \badgeParams{400M-7B} \badgeDate{02/2025}\quad \citet{tschannen2025siglip2multilingualvisionlanguage} extended SigLIP with an autoregressive decoder for generative tasks while maintaining the dual-encoder architecture. The vision encoder incorporates aspect ratio preservation \citep{navit_dehghani_2023}, variable sequence lengths following FlexiViT \citep{Beyer_2023_CVPR}, and a local-to-global consistency loss \citep{naeem2023silcimprovingvisionlanguage}. The consistency loss uses self-distillation, where the model learns from partial image views to match full-view representations, improving performance on both vision-language and computer vision tasks.

\paragraph{Pali-X.} \badgeEncoderDecoder{}~\badgeParams{55B}~\badgeDate{05/2023}\quad Pali-X \citep{chen2023palix} is based on the 32-billion parameter UL2 encoder-decoder model \citep{tay2023ul2unifyinglanguagelearning} with a T5-based architecture \citep{raffel2023exploringlimitstransferlearning}. The vision part is a Vision Transformer with 22B parameters \citep{dehghani2023scaling}, using patch embeddings directly as input in a one-tower configuration. Pre-training employs multiple training objectives: incorporating traditional VL tasks such as image captioning, optical character recognition, and object detection, as well as span corruption, similar to the text-only T5 training. The model goes beyond static images and also enables video processing.

% \JL{What about MAGMA, didn't they do the same thing earlier?}
% MAGMA is based on GPT-J (zero-shot, while only the CLIP-ViT is trained), and it hasn't which hasn't seen any multilingual data 
% Also, isn't Vicuna based on the first LLaMA? - yes you are right here

\paragraph{LLaVA.} \badgeDecoder{}~\badgeParams{7B-13B}~\badgeDate{04/2023}\quad LLaVA \citep{liu2023visualinst} was one of the first models to add vision capabilities to an instruction-tuned LLM. It combines Vicuna \citep{vicuna2023}, a decoder-only model derived from the first version of LLaMA \citep{touvron2023llamaopenefficientfoundation}, and the CLIP-ViT visual encoder. The architecture processes visual patches and text tokens within a unified stream. While the instruction tuning data is primarily in English, the underlying LLaMA weights incorporate multilingual knowledge from training on the ShareGPT4V dataset \citep{chen2023sharegpt4vimprovinglargemultimodal}, which includes content in Chinese and Japanese. The subsequent LLaVA 1.5 release \citep{liu2024improvedbaselinesvisualinstruction} refined the visual encoder and expanded both the pre-training and evaluation datasets but did not cover more languages.

\paragraph{mBLIP.} \badgeDecoder{}~\badgeParams{3B-13B}~\badgeDate{07/2024}\quad \citet{geigle-etal-2024-mblip} re-aligned BLIP-2's \citep{li2023blip2} image encoder to multilingual LLMs (mT0 and BLOOMZ - \citet{muennighoff-etal-2023-crosslingual}) using machine-translated training data across 95 languages. The architecture combines a Vision Transformer, Q-Former, and a multilingual language model. Training updates only the Q-Former, projections, and query tokens, with LoRA \citep{he_li_zhang_yang_wang_2023} for parameter-efficient decoder adaptation. The model shows competitive performance on IGLUE and XM3600 benchmarks, significantly outperforming English-only Vision-LLMs such as LLaVA 1.5, though later models surpass mBLIP on text-generation tasks.

% \JL{Do we have a citation to show that it sucks in tasks that go beyond classification?}
% CVQA (multiple choice QA, so still classification but in 4 classes) paper https://arxiv.org/pdf/2406.05967
% In contrast, open models like InstructBLIP and mBLIP-mT0 exhibit lower performance, particularly in local language prompts, indicating a need for more diverse training data and refined fine-tuning processes
% in Pangea chat benchmarks, the performance is low but the authors do not mention it in a sentence

\paragraph{Pangea.} \badgeDecoder{}~\badgeParams{7B}~\badgeDate{10/2024}\quad \citet{yue2024pangeafullyopenmultilingual} trained Pangea on a large multilingual dataset spanning vision-language tasks and multilingual document understanding (40\% English, 60\% other languages). The model offers two decoder-only variants using CLIP-ViT features: one based on Vicuna 1.5 \citep{vicuna2023} and another on Qwen 2 \citep{yang2024qwen2technicalreport}. Pangea outperforms LLaVA on several multilingual benchmarks, including MARVL and CVQA.

\paragraph{PALO.} \badgeDecoder{}~\badgeParams{1.7B-13B}~\badgeDate{02/2024}\quad \citet{maaz2024palopolyglotlargemultimodal} introduced two PALO variants: a 13B parameter version based on LLaVA with CLIP-ViT visual embeddings, and a 1.7B parameter MobilePALO \citep{chu2023mobilevlmfaststrong} using depth-wise separable convolutions for efficient visual token downsampling. Pre-training updates only the visual projection modules on CC-594k, followed by instruction tuning on machine-translated LLaVA-Bench \citep{liu2023visualinst} conversations. The authors report balanced performance across ten languages but do not provide detailed evaluation metrics.

\paragraph{Maya.} \badgeDecoder{}~\badgeParams{8B}~\badgeDate{12/2024}\quad \citet{alam2024mayainstructionfinetunedmultilingual} extended the Aya-23 language model \citep{aryabumi2024aya23openweight} to multimodal settings using SigLIP \citep{zhai2023sigmoidlosslanguageimage} visual features. The model is trained on toxicity-filtered and machine-translated LLaVA data across eight languages. Despite multilingual training, the evaluation focuses only on English multimodal benchmarks.

\paragraph{Centurio.} \badgeDecoder{}~\badgeParams{8B-72B}~\badgeDate{01/2025}\quad
\citet{geigle2025centuriodriversmultilingualability} investigated finetuning multilingual capabilities in VL models when using Aya and Qwen 2.5-VL. Balancing the data to 50\% English and 50\% other languages consistently improves performance across all language tiers \citep{joshi-etal-2020-state}. Centurio achieves competitive results with GPT-4 \citep{song-etal-2025-missing} on multilingual benchmarks. The study also addresses language fidelity issues, where models respond in incorrect languages, which have a significant impact on multilingual performance.

\paragraph{Phi-4-VL.} \badgeDecoder{} \badgeParams{3.8B} \badgeDate{03/2025}\quad
\citet{microsoft2025phi4minitechnicalreportcompact} introduced Phi-4-Mini, a compact 3.8B parameter multimodal LLM supporting text, images, and speech. The architecture uses grouped query attention (24 query heads, 8 key-value heads) to reduce KV cache consumption and LongRoPE \citep{ding2024longropeextendingllmcontext} for extended context windows of up to 2048k tokens. While the model officially supports multilingual speech across eight languages, multilingual image-text capabilities are not officially supported; however, though empirical evidence suggests that the model possesses some multilingual vision-language abilities.

\paragraph{The Qwen Model Family.}

Developed by Alibaba Cloud, the Qwen model family accompanies each language model release with corresponding VL versions.

\citet{wang2024qwen2vl} introduced Qwen 2-VL with Naive Dynamic Resolution for processing images at various resolutions and Multimodal Rotary Position Embedding (M-RoPE) for positional encoding across text, images, and videos. The family spans 2B to 72B parameters, with the largest version outperforming GPT-4o on diverse benchmarks, including multilingual OCR. 

\citet{bai2025qwen25vltechnicalreport} enhanced the architecture in Qwen 2.5-VL with a redesigned vision encoder using selective self-attention, windowed attention, RMSNorm \citep{zhang_biao2019_rmsnorm}, and SwiGLU activations \citep{pmlr-v70-dauphin17a}. Visual features utilize M-RoPE encoding with a group projector that clusters adjacent patches to reduce the token count. 

\citet{bai2025qwen3vltechnicalreport} further improved the model in Qwen 3-VL, introducing a thinking mode for multi-step reasoning via chain-of-thought finetuning and reinforcement learning. Pre-training expanded to 36T tokens across 119 languages, with architectures ranging from 0.6B to 235B parameters in both dense and mixture-of-experts variants. Detailed training data and language-specific metrics are not fully disclosed.

\paragraph{The LLaMA Model Family.} Developed by Meta with proprietary training data, the LLaMA family includes multimodal capabilities starting with version 3.2. 

\citet{grattafiori2024llama3herdmodels} combined a MetaCLIP-ViT image encoder \citep{xu2024demystifyingclipdata}, a vision projector, and a LLaMA decoder, trained in five stages, including multimodal training on 6 billion image-text pairs across multiple languages. Despite multilingual training data across 7 non-English languages, official support is restricted to English for image-text tasks, while speech processing supports 34 languages.

LLaMA 4 adopts a mixture-of-experts architecture.
\citep{jacobs1991adaptive,fedus2021switch} that routes multimodal streams through specialized feed-forward network sublayers: both shared and 15 specialized FFN experts are combined via residual connections. The system has 288 billion active parameters within a 2 trillion-parameter architecture, with distilled variants (Maverick and Scout) supporting up to 10 million token context lengths.

\paragraph{The Gemma Model Family.} Developed by Google as open-weight models with proprietary training data, the Gemma family includes several multimodal variants.

\citet{beyer2024paligemmaversatile3bvlm} introduced Pali-Gemma, trained on the proprietary WebLI dataset containing samples from over 100 languages. The model incorporates pre-training objectives, including multi-object detection and instance segmentation via Pix2Seq \citep{chen2022pix2seqlanguagemodelingframework}, which reformulates object localization as sequence generation. Performance on reasoning tasks like MARVL is competitive with Centurio.

\citet{gemmateam2025gemma3technicalreport} released Gemma 3 with Grouped Query Attention and a 5:1 ratio of local sliding-window to global self-attention, supporting 128k token contexts. The SigLIP vision encoder (896×896 resolution) uses Pan and Scan, an adaptive windowing algorithm that segments images into non-overlapping patches to handle non-square inputs. 

\citet{zhang2025t5gemma2seeingreading} developed T5Gemma 2, a lighter encoder-decoder variant initialized from Gemma 3 and trained on 2 trillion multilingual multimodal tokens using the UL2 objective \citep{tay2023ul2unifyinglanguagelearning}. The architecture introduces efficiency improvements through shared encoder-decoder embeddings and merged self- and cross-attention modules, though performance remains below that of the original T5 Gemma.

\section{Multilingual Vision-Language Benchmarks}\label{sec:benchmarks}

\begin{table*}
\caption{Overview of the recent multilingual multimodal benchmarks. The listed number of languages is followed by a tiny histogram showing the number of languages in each of the 5 tiers; the leftmost, dark gray bar represents the most-studied group (5), while the rightmost, light bars represent the least-studied languages (1). Then, we list the proportion of the world population covered by the benchmark and syntactic (MPSB) and morphological diversity (GBI) of the languages. For the number of texts and images, we indicate if they differ across languages ($\neq$) or whether they are the same/translations of each other ($=$).}

\centering\renewcommand{\arraystretch}{1.4}
\addtolength{\tabcolsep}{-2.0pt}
% \begin{tabular}{l cccccccc@{\hskip-2mm} c@{\hskip-4mm}}
\begin{tabular}{l ccccccccc}
\toprule
Dataset & citation & \multicolumn{2}{c}{\# languages} & \% pop. & MPSD & GBI & \# texts & \# images & focus \\ \midrule
MARVL & \tabcite{liu-etal-2021-visually} & 5 &  \tiers{0.2}{0.2}{0.4}{0.2}{0.0}{0.0} & 13.8 & 69.1 & 56.4 & $\neq$ 5.7K & $\neq$ 4.9K & culture \\
XVNLI & \tabcite{bugliarello2022} & 5 &  \tiers{0.8}{0.2}{0.0}{0.0}{0.0}{0.0} & 16.9 & 54.1 & 49.7 & $=$ 1.1K & $=$ 357 & semantics \\
xGQA   & \tabcite{nooralahzadeh2022} & 8  &  \tiers{0.29}{0.43}{0.29}{0.0}{0.0}{0.0} & 21.8 & 59.0 & 63.6 & $=$ 2.4M & $=$ 85K & semantics \\
xFlickrCo  & \tabcite{bugliarello2022} & 8  &  \tiers{0.62}{0.25}{0.12}{0.0}{0.0}{0.0} & 27.5 & 58.2 & 65.1 & $=$ 2K & $=$ 2K & semantics \\
WIT   & \tabcite{srinivasan2021} & 108  &  \tiers{0.27}{0.27}{0.45}{0.0}{0.0}{0.0} & 58.9 & 67.1 & 96.4 & $\neq$ 9.6k & $\neq$ 6.2k & locations \\
Multi30K & \tabcite{barrault-etal-2018-findings} & 4  &  \tiers{0.75}{0.25}{0.0}{0.0}{0.0}{0.0} & 6.6 & 53.3 & 33.8 & $=$ 620K & $=$ 31K & semantics \\
XM3600 & \tabcite{thapliyal-etal-2022-crossmodal} & 36 &  \tiers{0.19}{0.42}{0.25}{0.03}{0.11}{0.0} & 49.2 & 62.9 & 91.8 & $\neq$ 261K & $=$ 3.6K & culture \\
MaXM   & \tabcite{changpinyo-etal-2023-maxm} & 7 & \tiers{0.43}{0.14}{0.43}{0.0}{0.0}{0.0} & 21.5 & 56.7 & 64.6 & $\neq$ 2K & $=$ 0.3K & semantics \\
XTD   & \tabcite{aggarwal2020} & 7 & \tiers{0.29}{0.71}{0.0}{0.0}{0.0}{0.0} & 22.1 & 56.9 & 71.3 & $=$ 7K & $=$ 1K & semantics \\
CoMMuTE  & \tabcite{futeral-etal-2023-tackling} & 6  &  \tiers{0.67}{0.33}{0.00}{0.00}{0.00}{0.0} & 18.6 & 60.3 & 66.2 & $\neq$ 1.9K & $=$ 372 & ambiguity \\
ArtElingo-28  & \tabcite{mohamed-etal-2024-culture} & 28  &  \tiers{0.04}{0.16}{0.32}{0.24}{0.24}{0.08} & 17.2 & 72.0 & 83.6 & $\neq$ 200K & $=$ 2K & \focuscell{emotional \\ subjectivity} \\
M5 \scriptsize(VGR, VLOD) & \tabcite{schneider-sitaram-2024-m5} & 12  &  \tiers{0.17}{0.17}{0.25}{0.33}{0.08}{0.08} & 16.8 & 61.7* & 74.9 & $\neq$ 1.4K & $\neq$ 6.8K & culture \\
% Maya & \tabcite{alam2024mayainstructionfinetunedmultilingual} & 8 &  \tiers{0.75}{0.25}{0.00}{0.00}{0.00}{0.0} & 33.7 & 59.7 & 75.9 & $=$ 4.4M & $=$ 558K & \focuscell{semantics \\ (toxicity free)} \\
SMPQA & \tabcite{geigle2025centuriodriversmultilingualability} & 11 &  \tiers{0.36}{0.36}{0.18}{0.09}{0.00}{0.0} & 28.9 & 61.3 & 76.9 & $=$ 7K & $=$ 50 & synthetic \\
M3Exam & \tabcite{zhang2023m3exammultilingualmultimodalmultilevel} & 9 & \tiers{0.22}{0.33}{0.22}{0.11}{0.11}{0.0} & 21.4 & 64.6 & 65.1 & $\neq$ 2.8K & $\neq$ 3.1K & \focuscell{general \\ knowledge} \\
EXAMS-V & \tabcite{das-etal-2024-exams} & 11 & \tiers{0.45}{0.45}{0.09}{0.0}{0.0}{0.0} & 22.6 & 61.3 & 76.4 & $\neq$ 5K & $\neq$ 1.2K & \focuscell{general \\ knowledge} \\
WorldCuisines & \tabcite{winata-etal-2025-worldcuisines} & 24 & \tiers{0.25}{0.21}{0.17}{0.08}{0.21}{0.08} & 44.8 & 65.5* & 88.2 & $=$ 1M & $\neq$ 6K & culture \\
MLMemes & \tabcite{dimitrov-etal-2024-semeval} & 4 & \tiers{0.5}{0.0}{0.25}{0.0}{0.25}{0.0} & 8.2 & 59.5 & 43.1 & $\neq$ 25.3K & $\neq$ 10.8K & \focuscell{persuation \\ techniques} \\
%M-LLavaBench    & \tabcite{maaz2024palopolyglotlargemultimodal} &  10 & \tiers{0.6}{0.2}{0.2}{0.0}{0.0}{0.0}      & 38.2 & 61.8 & 75.9 & $=$ 2.1M & $=$ 1.3M & conversations \\
xMMMU   & \tabcite{yue2024pangeafullyopenmultilingual} &   7 & \tiers{0.57}{0.29}{0.14}{0.0}{0.0}{0.0} & 18.0 & 62.0 & 69.2 & $=$ 2.6K & $=$ 300 & semantics \\
% xChatBench & \tabcite{yue2024pangeafullyopenmultilingual} & 8 & \tiers{0.5}{0.25}{0.12}{0.0}{0.12}{0.0} & 28.9 & 63.3 & 63.1 & $=$ 400 & $=$ 50 & conversations \\
MTVQA   & \tabcite{tang2024mtvqabenchmarkingmultilingualtextcentric} &  10 & \tiers{0.5}{0.4}{0.1}{0.0}{0.0}{0.0}      & 16.4 & 62.0 & 77.9 & $\neq$ 28.6K & $\neq$ 8.7K & semantics \\
CVQA    & \tabcite{romero2024cvqaculturallydiversemultilingualvisual} &  31 & \tiers{0.1}{0.13}{0.29}{0.13}{0.32}{0.03} & 40.9 & 67.7 & 88.2 & $\neq$ 10.3K & $\neq$ 5.2K & culture \\
MVL-SIB & \tabcite{schmidt2025mvlsibmassivelymultilingualvisionlanguage} & 199 & \tiers{0.04}{0.1}{0.14}{0.08}{0.43}{0.23} & 70.2 & 70.8 & 95.4 & $=$ 3.1M & $=$ 70 & semantics \\
\begin{minipage}{2cm}KnowRecall \& \\ VisRecall\end{minipage} & \tabcite{wang2025travelinglanguagesbenchmarkingcrosslingual} & 15 & \tiers{0.5}{0.36}{0.14}{0.0}{0.0}{0.0} & 35.3 & 59.0 & 77.9 & $=$ 45K & $=$ 1.5K & culture \\
ALM-bench & \tabcite{vayani2025languagesmatterevaluatinglmms} & 99 & \tiers{0.06}{0.18}{0.28}{0.11}{0.32}{0.04} & 59.9 & 68.5 & 93.8 & $\neq$ 22K & $\neq$ 3K & culture \\

\bottomrule
\end{tabular}
\label{tab_mm_ml_benchmarks} 
\end{table*}

Comprehensive and responsible evaluation of multilingual vision-language models requires benchmarks that assess performance across diverse languages and cultural contexts. Unlike monolingual evaluation, multilingual benchmarks must navigate the fundamental tension between language neutrality and cultural awareness discussed in Section~\ref{sec:goals}. This challenge is implicitly reflected in the approaches taken by existing datasets, ranging from direct translation of English content to culturally grounded, independently developed evaluations.

We collect and analyze 23 recent multilingual vision-language benchmarks, summarized in Table~\ref{tab_mm_ml_benchmarks}. These datasets encompass a range of tasks, including visual reasoning, image-text retrieval, visual question answering, and image captioning.

For every dataset, we report the number of languages it covers and additional statistics on language coverage.
We categorize languages according to the resourcefulness tiers from \citet{joshi-etal-2020-state}, ranging from tier 5 (high-resourced) to tier 0 (left-behind), and visualize the distribution in the table.

To assess the linguistic diversity of each benchmark, we use two metrics introduced by \citet{ploeger-etal-2024-typological}: Mean Pairwise Syntactic Distance (MPSD), which computes the average syntactic distance between all possible language pairs in a dataset based on the features from the URIEL database \citep{littell-etal-2017-uriel}, and GramBank Feature Value Inclusion (GBI: \citealp{skirgard2023grambank}), which measures the proportion of all possible grammatical feature values from the GramBank database that are represented by the languages in the dataset. We are not aware of any metrics quantifying cultural diversity. However, we believe that linguistic diversity can also serve as a proxy for cultural diversity to some extent.

%\footnote{There are some limitations: 1) The use of WALS (World Atlas of Language Structures) language codes can occasionally be inconsistent with the ISO-639-3 standard, and 2) The lang2vec database does not include all languages. As a result, the value for benchmarks with very low resource languages might be inaccurate.}
%
Furthermore, we collect the number of native speakers (L1) from Wikidata\footnote{https://www.wikidata.org/} for each language and estimate the percentage of the world's population that is covered by the dataset.

We also report the size of the datasets, which is, however, difficult to compare directly. An image can be associated with multiple text samples and vice versa.
The number of texts in the table is the total number of samples from all languages. Samples also differ in granularity, ranging from sentences to paragraphs. For each dataset, we report the number of unique images. To show whether the texts are translations of each other and if the images are the same, we add symbols $=$ (the same across languages) and $\neq$ (different across languages).

Finally, we aim to summarize the main focus of each dataset using a single keyword.

\subsection{General Trends and Observations}

\paragraph{Language Coverage.} The benchmarks vary in both the number of tested languages and typological diversity. While some datasets, such as MVL-SIB \citep{schmidt2025mvlsibmassivelymultilingualvisionlanguage}, cover 199 languages, most benchmarks focus on 4 to 30 languages. Most datasets cover less than one-third of the world's population, whereas the 199-language MVL-SIB covers 70\% of the population; however, it contains only 70 images, which are not representative of the world as a whole.

Most benchmarks focus on rather high-resource languages, with tier 4--5 languages dominating the distributions shown in Table~\ref{tab_mm_ml_benchmarks}. Notable exceptions include ArtElingo-28 \citep{mohamed-etal-2024-culture} and MVL-SIB, which include a substantial number of low-resource languages (tiers 1-2).

Finally, there are a handful of benchmarks that claim to cover many cultures across continents, but the actual content is only visually diverse, while the textual component is represented only in English \citep{nayak2024benchmarkingvisionlanguagemodels,nwatu2025cultureaffordanceatlasreconciling,irawan2025visionlanguagemodelsconfused}.

\paragraph{Linguistic Diversity.}

High-diversity datasets such as ArtElingo-28 (MPSD: 72.0) and MARVL (MPSD: 69.1) prioritize typological diversity over simple language count, while larger collections like XM3600 have a higher number of languages; however, they are much less diverse (36 languages, MPSD: 62.9). The highest morphological diversity is achieved by WIT \citep{srinivasan2021} (GBI: 96.4) and MVL-SIB (GBI: 95.4), both based on data extracted from Wikipedia.

\paragraph{Task Coverage and Scale.}

The most commonly used monolingual evaluation benchmarks have their multilingual counterparts.
Visual reasoning tasks, typically formulated as classification, dominate, with datasets such as MARVL, M5-VGR \citep{schneider-sitaram-2024-m5}, and xGQA \citep{nooralahzadeh2022} testing semantic understanding through logical inference. Retrieval benchmarks such as xFlickrCo \citep{bugliarello2022} and XTD focus on cross-modal similarity assessment, while generative tasks are represented by image captioning datasets like XM3600. The dataset size varies dramatically, ranging from synthetic datasets such as SMPQA \citep{geigle2025centuriodriversmultilingualability} (50 images) to larger collections such as xGQA (85k images).

\paragraph{Evaluation Paradigms.}

The benchmarks can be categorized into three primary evaluation paradigms based on their approach to multilinguality. Translation-based datasets such as XVNLI \citep{bugliarello2022}, xGQA, and Multi30k \citep{barrault-etal-2018-findings} maintain identical visual content while translating textual components, enabling direct cross-lingual performance comparisons and supporting language-neutral evaluation. Culture-aware datasets like MARVL, ArtElingo-28, and WorldCuisines \citep{winata-etal-2025-worldcuisines} use region-specific images and culturally-grounded content, explicitly testing cultural awareness capabilities. 
% Other datasets, such as XM3600 (images from around the world with independent captions in different languages) and CVQA (local images with native-language questions and English translations), aim to balance semantic consistency with cultural authenticity. \JL{I don't understand what the last sentence means. If you don't either, we can delete it.}

\paragraph{Cultural vs. Semantic Focus.}

Roughly two-thirds of the benchmarks primarily evaluate semantic consistency and language neutrality, following translation-based methodologies that enable controlled cross-lingual evaluation. For classification tasks, it allows direct comparison between languages.

The rest, at least to some extent, evaluates cultural awareness through diverse imagery, subjective evaluation criteria, or region-specific content.
Culturally aware benchmarks typically use both distinct images and distinct texts across languages. We have not found a dataset with texts that are translations of each other but contain language- or region-dependent images. WorldCuisines slightly matches the condition: even if the VQA set contains only one image per dish, the collection includes an additional knowledge base with pictures of food from a specific country. However, there is no direct link between the two datasets, only a possible way to augment the first.
We observe an increased interest in culture-aware evaluation starting in 2024, so this distribution might change in the future.

\paragraph{Aggregated Benchmarks.}

Several recent publications have introduced benchmark collections that aggregate previously released datasets, creating comprehensive evaluation suites. Notable examples include IGLUE \citep{bugliarello2022} and M5 \citep{schneider-sitaram-2024-m5}, both of which incorporate the MARVL reasoning dataset \citep{liu-etal-2021-visually}. While these collections aim to provide standardized evaluation frameworks, they might create confusion about dataset novelty and evaluation scope. In this survey, we list only the original contributions of each collection, rather than benchmarks reused from previous work.

\paragraph{Methodological Limitations.}

Current benchmarks have several systemic limitations.
First, most datasets rely heavily on English as an intermediate language, either through direct translation or English-centric image collections.
Second, visual content often originates from mostly Western sources (e.g., Flickr30k \citet{young-etal-2014-image}), potentially introducing cultural bias. Moreover, many of the images are more than ten years old and thus might not reflect, for instance, how fashion and many technological artifacts look today.
Third, evaluation metrics typically assume semantic equivalence across languages, which may not capture culturally dependent differences in interpretation.
Finally, the scarcity of large-scale, culturally authentic datasets limits the comprehensive evaluation of cultural awareness capabilities in vision-language models.

\subsection{Detailed Benchmark Overview}

This section comments on the details of the benchmarks presented in Table~\ref{tab_mm_ml_benchmarks} that we summarized in the previous section. They are listed roughly in chronological order.

\paragraph{MARVL.}

The Multicultural and Multilingual Visual Language (MARVL) dataset, by \citet{liu-etal-2021-visually}, evaluates visual language reasoning, i.e., the classification of the logical agreement between an image and a sentence. Unlike other datasets translated from original English datasets, MARVL was independently developed for five languages, with a focus on cultural awareness: Turkish, Swahili, Chinese, Indonesian, and Tamil. The dataset consists of image-text pairs where, given two ordered images and a textual hypothesis, models must determine whether the hypothesis is true or false. 

\paragraph{XVNLI.}

Part of the IGLUE collection \citep{bugliarello2022}, the Cross-lingual Visual Natural Language Inference (XVNLI) dataset extends the task of natural language inference (NLI) into the multimodal, multilingual domain. The dataset builds upon the Stanford Natural Language Inference (SNLI) corpus \citep{bowman-etal-2015-large} and its multimodal version \citep{xie2019visual}. XVNLI \citep{agic-schluter-2018-baselines} covers 4 additional languages (Arabic, Spanish, French, and Russian) by translating the original English sentences. The task involves classifying the relationships between an image and two textual hypotheses as entailment, neutral, or contradiction. One sample is depicted in Figure~\ref{fig:xvnli_sample}.

\paragraph{xGQA.}

Derived from the English-only GQA dataset \citep{hudson_2019_CVPR}, xGQA introduces a multilingual visual question answering benchmark. The test set has been manually translated into seven mostly high-resource languages. It contains visually grounded questions that models must answer using a predefined set of approximately 2000 labels. Unlike most other IGLUE subsets, xGQA has a relatively large training set, totaling 2.4 million sentences and 85,000 unique images.

\paragraph{xFlickrCo.}

The xFlickrCo dataset, also a part of the IGLUE benchmark \citep{bugliarello2022}, is an evaluation set designed for cross-lingual image-text retrieval. It consists of 2k images, equally sourced from the Flickr30k dataset \citep{young-etal-2014-image} and the COCO \textit{Karpathy split} \citep{lin2014COCO, karpathy_2015_CVPR}. 

\paragraph{WIT.}

The Wikipedia-based Image-Text (WIT) dataset \citep{srinivasan2021} contains images extracted from Wikipedia articles, along with their corresponding captions extracted from the articles' abstracts, curated for vision-language retrieval tasks. The IGLUE version of WIT contains only a subset of the original dataset, with 11 out of the 108 original languages. Many captions on Wikipedia are originally in local languages, and many are translations. It is therefore not entirely clear to what extent this benchmark measures cultural awareness, especially given Wikipedia's bias toward universal and institutionalized knowledge.

\paragraph{Multi30k.}

The Multi30k dataset was originally developed for multimodal translation between English and German \citep{elliott-etal-2016-multi30k} and was later extended to English-French \citep{elliott-etal-2017-findings} and English-Czech \citep{barrault-etal-2018-findings}. Multimodal translation is the translation of image captions, where the visual information can be used to disambiguate the translation. Later, it was also repurposed for text-image retrieval.

\paragraph{XM3600}

\citet{thapliyal-etal-2022-crossmodal} built XM3600 for image captioning. For each of the 36 languages, 100 images were selected from The Open Images Dataset \citep{kuznetsova2020open}, which consists of open-domain images, primarily from Flickr. The authors then recruited bilingual annotators to provide captions in both English and their native languages for each image. The 36 languages were selected to ensure high morphological diversity (GBI $=91.8\%$). 

\paragraph{MaXM.}

\citet{changpinyo-etal-2023-maxm} created a visual open-ended question-answering dataset using a subset of image-caption pairs from XM3600. The questions, answers, and their translations across languages were generated automatically and then manually validated by annotators.

\begin{figure}
    \centering
    \includegraphics[width=\linewidth]{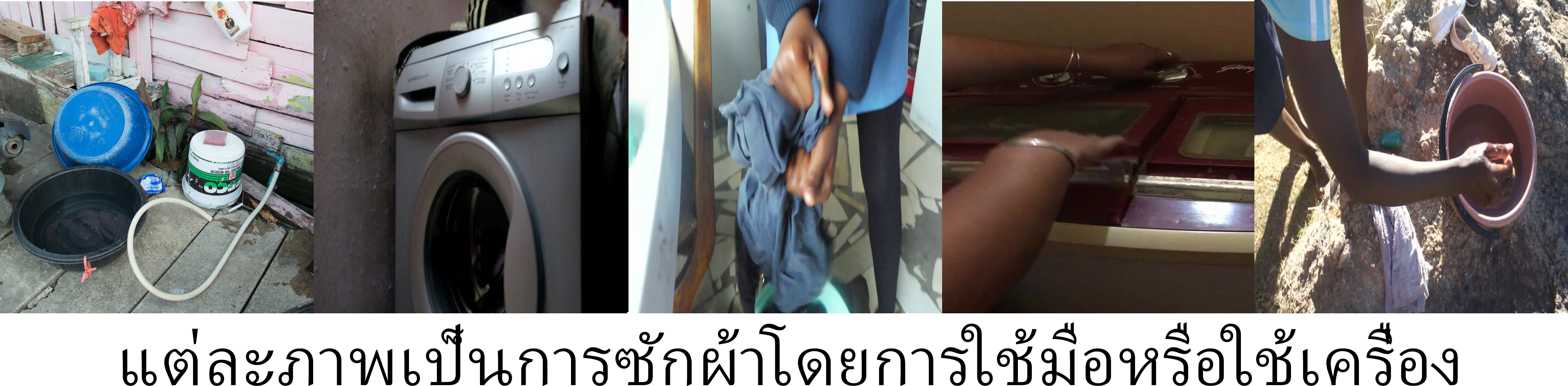}
    \caption{Visual-language outlier detection example from M5-VLOD set. The original Thai caption can be translated as: "Each photo is of washing clothes by hand or using a machine". The outlier is 1, the second image with the idle washing machine.}
    \label{fig:m5_vlod_sample}
\end{figure}

\paragraph{XTD.}

\citet{aggarwal2020} add more languages to the MSCOCO2014 test set \citep{rajendran-etal-2016-bridge}, one of the most frequently used datasets in computer vision, which has a rich set of annotations. XTD contains 1,000 image-text pairs, with human annotators translating them into 9 languages. Additionally, they included Japanese captions from the STAIR dataset \citep{yoshikawa-etal-2017-stair}, which refer to the same images. The evaluation task is multilingual image-text retrieval.

\paragraph{CoMMuTE.}

This dataset \citep{futeral-etal-2023-tackling} targets lexical ambiguity in multimodal machine translation from English to Czech, German, and French. It consists of 155 English short ambiguous sentences. Each sentence is accompanied by two images that provide distinct contexts, leading to distinct translations in the target language. %In the published experiments, the models are evaluated strictly on machine translation quality in both contexts. However, we believe we can devise a classification task that asks the model to choose the best visual context for a randomly selected pair of sentences.

\paragraph{ArtElingo-28.}

\citet{mohamed-etal-2024-culture} introduces a unique dataset of (mostly Western) paintings with captions in 28 languages, each individual pair being associated with one of 9 emotion labels. With languages mostly from Africa and Southeast Asia, ArtElingo-28 achieves high linguistic diversity (MPSD $72.0$). The main task is emotion classification based on the subjective emotions embedded in culturally dependent captions. This dataset offers an unusual perspective on cultural awareness: rather than focusing on local imagery, it captures diverse perspectives on cultural artifacts across cultures.

%However, several LLMs, such as mBLIP and MiniGPT4, were also tested in multilingual image captioning and evaluated with machine translation metrics.

\paragraph{M5-VGR \& MG-VLOD.}

M5 \citep{schneider-sitaram-2024-m5} is a collection of existing datasets, plus two newly introduced datasets: M5-VGR for vision-grounded reasoning and M5-VLOD for vision-language outlier detection. Both tasks cover 13 languages, including very low-resource languages such as Berber. M5-VGR follows MARVL's reasoning format. The task in M5-VLOD is to identify which image out of five similar images does not correspond to a given prompt. Figure~\ref{fig:m5_vgr_sample} and Figure~\ref{fig:m5_vlod_sample} show one example from each subset. 

\paragraph{Multilingual Memes.}

\citet{dimitrov-etal-2024-semeval} created a dataset for the 2024 SemEval shared task on classifying persuasion techniques in memes. The dataset covers a few languages (Bulgarian, Northern Macedonian, Arabic, and English) and is only available upon request. Memes are contemporary cultural artifacts full of irony and sarcasm that differ across cultures. These datasets thus test cultural awareness from an unusual and interesting perspective.

\paragraph{WorldCuisines.}

\citet{winata-etal-2025-worldcuisines} uses knowledge of food as a proxy for cultural knowledge. The WorldCuisines benchmark consists of food-related visual question answering across 24 languages, covering over 2,000 food concepts from almost all countries worldwide. In addition to QA, the dataset contains geographic coordinates for each food concept. Question-answer pairs were collected in English and manually translated.
%
%In the dish selection, the authors chose only those with a Wikipedia article and that have "a distinct cultural significance". Regarding the latter, we note that there are also several strong culinary similarities among neighboring countries and communities.

\paragraph{xMMMU.}

This dataset, published together with the Pangea model \citep{yue2024pangeafullyopenmultilingual} is an extended version of the English-language MMMU \citep{yue2023mmmu} benchmark generated via machine translation. It only contains 300 questions from the validation split of MMMU  translated into 6 languages using GPT-4o. It mostly covers knowledge from college exams. Although it is rather small and covers only a few languages, the performance on this dataset was reported in technical reports for several commercial models, such as Gemma 3, LLaMa 3, and Qwen 3-VL.

\begin{figure}
    \centering
    \includegraphics[width=0.7\linewidth]{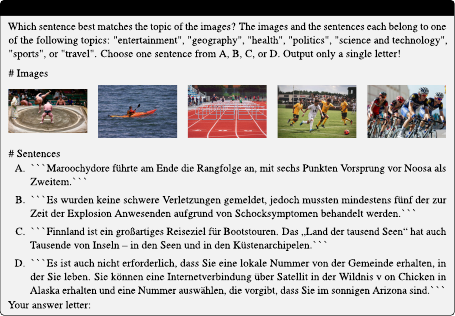}
    \caption{Sample from MVL-SIB input, copied from the original publication, under CC BY-SA 4.0.}
    \label{fig:mvl_sib_sample}
\end{figure}

\paragraph{MTVQA.}

MTVQA \citep{tang2024mtvqabenchmarkingmultilingualtextcentric} focuses on text-rich image understanding across 10 high-resource languages. It contains 8.7k images with multilingual text, extracted from real-life objects, e.g., grocery products with multilingual instructions, bills, books, documents, etc. The dataset contains 28.6k question-answer pairs regarding both text content and metadata (e.g., font family, font color, etc). Unlike most QA benchmarks, this benchmark evaluates the open-ended generation of short answers. The evaluation, however, is quite simplistic, using exact-match accuracy with a single correct answer.

\paragraph{CVQA.}

\begin{figure}
    \centering
    \includegraphics[width=0.9\linewidth]{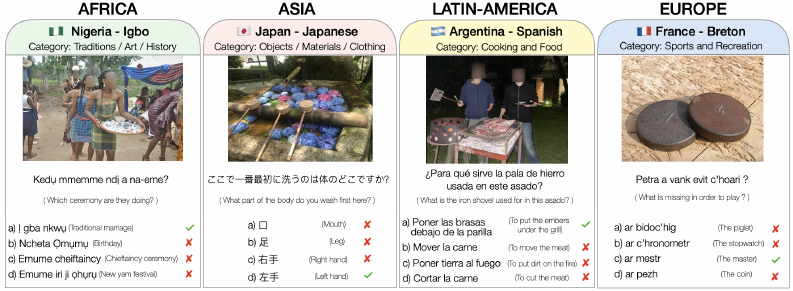}
    \caption{Sample from CVQA dataset, copied from the original publication, under CC BY-SA 4.0.}
    \label{fig:cvqa_sample}
\end{figure}

CVQA \citep{romero2024cvqaculturallydiversemultilingualvisual} targets cultural diversity through multiple-choice visual question answering across 39 language-country pairs. Each sample is in both the local language and English. We show one sample from the original publication in Figure~\ref{fig:cvqa_sample}. The topics are classified into 10 categories, such as food, vehicles, and fashion. We consider this to be one of the most comprehensive benchmarks for cultural awareness. It uses authentic imagery from different countries and a comprehensive set of cultural categories. The only limitation is that it is designed to be evaluated using accuracy in a multiple-choice setup, which does not reflect how generative models are usually used.

\paragraph{MVL-SIB.}

\citet{schmidt2025mvlsibmassivelymultilingualvisionlanguage} achieves the highest language coverage and typological diversity, with 200 languages from the FLORES dataset, while using only 70 unique images. The proposed task, cross-modal topical matching, considers a set of five images and five captions. The model must choose the caption that best matches the standalone set of images. The language coverage makes it a good resource for comparing languages. However, with such a small number of languages, the visual part of the dataset cannot be representative of all the languages. Figure~\ref{fig:mvl_sib_sample} contains one sample from the original article.

\paragraph{KnowRecall and VisRecall.}

\citet{wang2025travelinglanguagesbenchmarkingcrosslingual} proposed two datasets focused on landmarks for evaluating cross-lingual consistency. KnowRecall is a multiple-choice VQA dataset, and VisRecall focuses on image captioning. The recall keyword stems from a hypothesis that the models have encountered the mentioned landmarks during pre-training. Since a landmark is always located in a country or region with a local language, the authors evaluated the models' performance in English, the local language, and the average score across all other languages (global language). The global score is always lower than that of English and the local language. At the same time, English performance exceeds local performance in most cases, but by a small margin. The article is controversial because it was recently withdrawn from arXiv for academic misconduct. However, we still consider it an important resource.

\paragraph{ALM-bench.}

ALM-bench \citep{vayani2025languagesmatterevaluatinglmms} covers several domains (e.g., heritage, customs, architecture) and multiple question formats (e.g., yes/no, multiple-choice, and open-ended). The dataset comprises approximately 22K samples, which is relatively smaller than its predecessors, and includes a detailed error analysis for each domain. The evaluators labeled the possible reasons for the best model's incorrect answers, with the most frequent reasons being lack of knowledge, reasoning errors, perception errors, language errors, translation errors, and lack of cultural understanding.

\paragraph{Drawing/scheme/exam datasets.}

Our table includes three additional multilingual sets focused more on drawings, charts, and plots than on scene images. \emph{SMPQA} is a synthetic dataset with only 50 images, generated from a pattern of rules by \citet{geigle2025centuriodriversmultilingualability}, which can be easily translated. \emph{M3Exam} \citep{zhang2023m3exammultilingualmultimodalmultilevel} and \textbf{EXAMS-V} \citep{das-etal-2024-exams} consist of collected exam questions, both text-only and multimodal, from different parts of the globe. We reported only the multimodal samples in Table~\ref{tab_mm_ml_benchmarks}.

% \paragraph{SMPQA} SMPQA is a synthetic dataset with only 50 images, generated from a pattern of rules by \citet{geigle2025centuriodriversmultilingualability}, which can be easily translated. The images are mostly statistical plots, while the questions aim to test (a) the multilingual OCR capability of the models or (b) ground the text from the input questions to the ones written inside the plots.

% \paragraph{M3Exam.} \citet{zhang2023m3exammultilingualmultimodalmultilevel} collected exam questions, both text-only and multimodal, from different parts of the globe and published them as M3Exam. Only $23\%$ of the dataset consists of image-dependent questions; therefore, the number of samples we report in Table~\ref{tab_mm_ml_benchmarks} refers only to this subset. 

% \paragraph{EXAMS-V.} Similar to M3Exam, \citet{das-etal-2024-exams} collected multi-disciplinary questions in a few more languages, coming up with more multimodal samples. For our statistics, we consider only the samples with figures. 

% \JL{I would take the three datasets above and merge them into a short paragraph at the end of this section, saying that they do not focus on scene images, but rather writing and drawing.}

\section{Future Research Directions}

Our survey includes patterns in architectural design, training methodologies, and evaluation practices. This section synthesizes these observations and identifies directions for future research.

\paragraph{Low-Resource Language Support.}

Evaluation coverage reveals a significant gap between claimed and verified multilingual capabilities. Most benchmarks test only 4--30 languages, while models claim support for 100+, leaving cross-lingual transfer empirically unverified for the majority of languages. Only 5 benchmarks include tier 0-1 languages (ArtElingo-28, M5, CVQA, MVL-SIB, EXAMS-V), and most models show minimal evaluation for languages below tier 3. Model scaling from hundreds of millions to over two trillion parameters has improved performance but increased computational requirements, creating accessibility challenges for low-resource languages, where computing infrastructure is often limited. Research might focus on cross-lingual transfer from high- to low-resource languages, e.g., via parameter-efficient fine-tuning methods that preserve pre-trained multilingual knowledge while adapting to new languages and leveraging external knowledge sources such as dictionaries or parallel Wikipedia.

\paragraph{Explicit Cultural Awareness Objectives.}

Training methodologies explicitly optimize for language neutrality through contrastive learning on parallel data (MURAL, NLLB-CLIP), cross-lingual alignment objectives (CCLM), and masked language modeling with multilingual encoders. These objectives are mathematically well-defined and directly optimizable. In contrast, cultural awareness lacks explicit training objectives. Current approaches rely on diverse training data (ALIGN's 1.8B pairs across 100+ languages, mBLIP's multilingual image-text pairs) without principled optimization for cultural variation. This asymmetry exists because language neutrality can be formalized through representation similarity metrics, while cultural awareness requires modeling context-dependent interpretation that varies across communities. Developing trainable objectives for cultural awareness remains an open problem. Promising directions include culture-conditioned generation that explicitly models regional variation in outputs, contrastive objectives that align culturally similar concepts while separating culturally distinct ones, or region-specific image collections with sufficient metadata.

\paragraph{Comprehensive Evaluation Frameworks.}

Current benchmarks have three main limitations. First, approximately two-thirds use classification tasks (MARVL, xGQA, XVNLI), which do not assess the open-ended generation capabilities of decoder-only models that now dominate the field. Second, as noted above, the evaluation covers far fewer languages than the models support. Third, generation metrics (BLEU, chrF) vary across languages and fail to capture cultural appropriateness. The field needs generation-focused benchmarks with human-evaluation protocols for cultural appropriateness, benchmarks covering more languages with balanced representation across language families and resource tiers, and diagnostics for systematic errors across linguistic features and cultural dimensions. We currently lag in automated metrics for cultural appropriateness that correlate with human judgments and in compositional evaluation that tests novel combinations of visual concepts and linguistic structures.

\paragraph{Data Transparency and Documentation Standards.}

Commercial models exacerbate evaluation challenges due to limited documentation. While technical reports describe architectures in detail, the composition of the training data and the language distribution remain largely undisclosed, making a systematic analysis of data coverage impossible. This opacity is particularly problematic for cultural awareness, which cannot be explicitly optimized through training objectives and arises primarily from the composition of training data. Future work should address several documentation challenges: establishing metrics to quantify language distribution and cultural diversity in training data, developing methods to detect and characterize bias in multilingual and multicultural contexts, and creating protocols for appropriate cultural representation that avoid stereotyping. Additionally, the field would benefit from community participation in benchmark creation, particularly from speakers of underrepresented languages, and from systematic audit procedures for models deployed across diverse linguistic communities. These improvements would enable a more rigorous assessment of model capabilities and limitations across languages and cultures.

\section{Conclusions}

This survey examines multilingual vision-language models and their evaluation benchmarks. It analyzes 33 models and 23 benchmarks released between 2020 and 2025 and discusses patterns in development timelines, training objectives, and evaluation practices.

Multilingual vision-language models lag behind their monolingual counterparts and language-only multilingual models. The field has moved from encoder-only architectures (14 models) optimized for classification to decoder-only models (14 models) capable of generative tasks, with parameter counts increasing from 100--400 million to over 2 trillion parameters.
However, the biggest and most successful models from commercial entities often suffer from transparency issues. Their documentation emphasizes architectural innovations but provides limited details on the composition of the training data and the language distribution. This undermines reproducibility and the assessment of actual multilingual capabilities.

We identify a fundamental tension between \emph{language neutrality} and \emph{cultural awareness} in multilingual vision-language modeling. Language neutrality aims to ensure consistent representations of objective concepts across languages, facilitating cross-lingual transfer. Cultural awareness emphasizes adaptation to linguistic and cultural contexts, recognizing that meaning varies across communities.

Current training methodologies often explicitly optimize for language neutrality through contrastive learning on translated captions, cross-lingual alignment objectives, and parallel data, especially with encoder models. However, there are no explicit cultural awareness objectives that we can optimize for, as cultural diversity must always come from diverse training data.

Approximately two-thirds of multilingual evaluation benchmarks prioritize language neutrality when using translation-based methodologies (XVNLI, xGQA, Multi30k), with identical visual content and translations of each other. However, we observe an increased interest in cultural awareness from 2024 onward, with benchmarks such as ArtElingo-28, WorldCuisines, and CVQA incorporating region-specific imagery and culturally grounded content.

Current evaluation benchmarks often rely on classification tasks, which do not adequately assess the generative capabilities of contemporary models. While classification enables straightforward, rigorous evaluation, it fails to capture the challenges of open-ended generation, cultural appropriateness, and linguistic authenticity across languages. Current language generation metrics often fail to accurately capture generation quality.

The tension between language neutrality and cultural awareness remains unresolved. While models demonstrate cross-lingual transfer capabilities, the emerging interest in cultural representation in evaluation benchmarks highlights gaps between evaluation goals and training methodologies.

\printbibliography

\end{document}